\documentclass[10pt, a4paper, logo]{googledeepmind} %

\usepackage{times}

\usepackage{hyperref}

\usepackage{amsmath,amsfonts,bm}
\usepackage{natbib}
\usepackage{fancyhdr}

\def\eqref#1{equation~\ref{#1}}

\def\1{\bm{1}}

\def\rr{{\textnormal{r}}}

\def\va{{\bm{a}}}

\def\vd{{\bm{d}}}

\def\vq{{\bm{q}}}

\def\vs{{\bm{s}}}

\def\vx{{\bm{x}}}
\def\vy{{\bm{y}}}

\def\vtau{{\bm{\tau}}}

\def\mD{{\bm{D}}}

\DeclareMathAlphabet{\mathsfit}{\encodingdefault}{\sfdefault}{m}{sl}
\SetMathAlphabet{\mathsfit}{bold}{\encodingdefault}{\sfdefault}{bx}{n}

\def\gA{{\mathcal{A}}}
\def\gB{{\mathcal{B}}}

\def\gQ{{\mathcal{Q}}}

\def\gX{{\mathcal{X}}}

\usepackage{hyperref}
\usepackage{url}
\usepackage{booktabs}
\usepackage{graphicx}
\usepackage{subcaption}
\usepackage{amssymb}
\usepackage[most]{tcolorbox}
\usepackage[table]{xcolor}
\tcbuselibrary{breakable} 
\usepackage{amsmath} 
\usepackage{siunitx}
\usepackage{colortbl}
\usepackage{makecell}
\usepackage{tcolorbox}
\definecolor{promptpurple}{RGB}{242, 242, 255} 
\definecolor{bordergray}{RGB}{80, 80, 80}
\definecolor{promptgreen}{RGB}{242, 255, 242} 
\definecolor{promptblue}{RGB}{242, 248, 255} 

\usepackage{microtype}
\usepackage{color}
\usepackage{wrapfig}
\usepackage{natbib}
\usepackage{colortbl}
\usepackage{microtype}
\usepackage{graphicx}
\usepackage{booktabs} 
\usepackage{array}
\usepackage{textcomp}
\usepackage{stfloats}
\usepackage{float}
\usepackage{verbatim}
\usepackage[ruled, linesnumbered, lined]{algorithm2e}
\usepackage{multirow}
\usepackage{enumitem}

\definecolor{BestColor}{HTML}{C8E6C9}  
\definecolor{SecondBestColor}{HTML}{FFF9C4} 

\usepackage{tcolorbox}
\usepackage{amsmath,amsfonts}

\definecolor{ggg}{RGB}{26,179,0}
\definecolor{rrr}{RGB}{179,0,0}
\definecolor{oodc}{RGB}{31,73,121}
\definecolor{idc}{RGB}{68,142,68}
\def\rr{\textcolor{red}}
\def\bb{\textcolor{blue}}
\definecolor{mygray}{gray}{0.9}

\def\Bias#1#2{\bm{b}}

\definecolor{gA}{RGB}{120,190,150} 
\definecolor{pB}{RGB}{170,140,210} 
\definecolor{bC}{RGB}{120,170,220} 
\definecolor{yD}{RGB}{230,210,120} 

\newtcolorbox{examplebox}[2][]{ 
    breakable, 
    enhanced, 
    colback=white, 
    colframe=cyan, 
    coltitle=white, 
    fonttitle=\bfseries, 
    title=#2, 
    overlay middle={\draw[cyan, line width=1pt](frame.south west)--(frame.south east);}, 
    overlay last={\draw[cyan, line width=1pt](frame.south west)--(frame.south east);}, 
    #1 
}

\usepackage[T1]{fontenc}
\usepackage{booktabs}      
\usepackage{graphicx}      
\usepackage{pifont}      
\usepackage[table]{xcolor} 
\usepackage{siunitx}       
\usepackage{etoolbox}      
\usepackage[normalem]{ulem}     
\usepackage{pgf} 

\definecolor{impcolor}{HTML}{2E8B57} 

\newcommand{\improvementstyle}[1]{$^{\textcolor{impcolor}{\tiny #1}}$}

\newcommand{\scoreimp}[2]{%
  \textbf{#1}%
  \ifstrequal{#2}{+0.0}{}{%
    \ifstrequal{#2}{0.0}{}{%
      \makebox[0pt][l]{\improvementstyle{#2}}%
    }%
  }%
}

\title{MARCH: {M}ulti-{A}gent {R}einforced Self-{C}heck for LLM {H}allucination}

\author[1,2]{Zhuo Li$^{*}$\textsuperscript{\dag}}
\author[1]{Yupeng Zhang$^{*}$}
\author[1]{Pengyu Cheng\textsuperscript{\ddag}}
\author[1]{Jiajun Song}
\author[1]{Mengyu Zhou}
\author[1]{Hao Li}
\author[1]{Shujie Hu}
\author[1]{Yu Qin}
\author[1]{Erchao Zhao}
\author[1]{Xiaoxi Jiang}
\author[1]{Guanjun Jiang}
\affil[1]{Qwen Large Model Application Team, Alibaba}
\affil[2]{The Chinese University of Hong Kong, Shenzhen}

\begin{abstract}
Hallucination remains a critical bottleneck for large language models (LLMs), undermining their reliability in real-world applications, especially in Retrieval-Augmented Generation (RAG) systems. 
While existing hallucination detection methods employ LLM-as-a-judge to verify LLM outputs against retrieved evidence, they suffer from inherent \textit{confirmation bias}, where the verifier inadvertently reproduces the errors of the original generation.
 To address this, we introduce \textbf{M}ulti-\textbf{A}gent \textbf{R}einforced self-\textbf{C}heck for \textbf{H}allucination (MARCH), a framework that enforces rigorous factual alignment by leveraging deliberate \textit{information asymmetry}.
MARCH orchestrates a collaborative pipeline of three specialized agents: a Solver, a Proposer, and a Checker.
 The Solver generates an initial RAG response, which the Proposer decomposes into claim-level verifiable atomic propositions. Crucially, the Checker validates these propositions against retrieved evidence in isolation, deprived of the Solver’s original output.
This well-crafted information asymmetry scheme breaks the cycle of self-confirmation bias. By training this pipeline with multi-agent reinforcement learning (MARL), we enable the agents to co-evolve and optimize factual adherence. Extensive experiments across hallucination benchmarks demonstrate that MARCH substantially reduces hallucination rates. Notably, an 8B-parameter LLM equipped with MARCH achieves performance competitive with powerful closed-source models. MARCH paves a scalable path for factual self-improvement of LLMs through co-evolution. The code is at \url{https://github.com/Qwen-Applications/MARCH}.
\end{abstract}

\begin{document}
\maketitle

\begin{figure*}[h]
    \centering
    \includegraphics[width=0.88\textwidth]{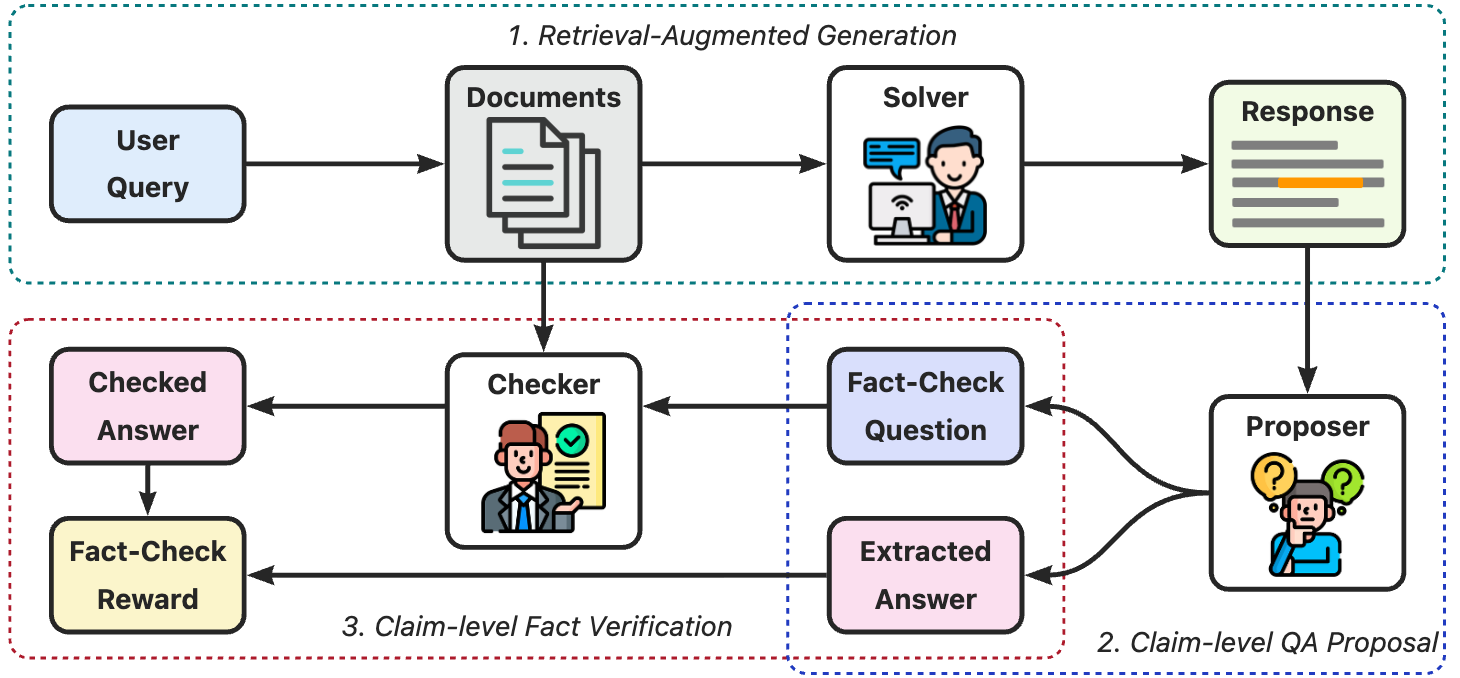}
\caption{Overview of the MARCH framework. The \textit{Solver} conducts retrieval-augmented generation base on the input query and related documents. The \textit{Proposer} then decomposes this response into atomic propositions (the orange line in the response block) and formulates verifiable question-answer pairs. The \textit{Checker} performs isolated verification by re-answering questions solely based on the retrieved evidence, without access to the \textit{Solver}’s original output. This information-asymmetric pipeline, where the policy model plays all three roles, is optimized via Multi-Agent Reinforcement Learning (MARL) to achieve robust factual alignment.
%
} 
    \label{fig:framework_1}
\end{figure*}

\section{Introduction}

Despite the remarkable success achieved by Large Language Models (LLMs)~\citep{openai2024gpt4technicalreport,touvron2023llamaopenefficientfoundation,yang2024qwen2,deepseekai2025deepseekr1incentivizingreasoningcapability,comanici2025gemini25pushingfrontier},
the persistence of \emph{hallucinations} remains a fundamental barrier to LLMs' trustworthy usage~\citep{kalai2025language,Huang_2025,ji2023survey,maynez2020faithfulnessfactualityabstractivesummarization}. Even enhanced within Retrieval-Augmented Generation (RAG)~\citep{lewis2020retrieval}, LLMs frequently generate responses that, while linguistically fluent, contradict the provided evidentiary context~\citep{xu2025hallucinationinevitableinnatelimitation}. Such context-conflicting hallucinations~\citep{liu2024lost} are particularly prevalent in data-intensive tasks, manifesting as inaccurate numbers, distorted temporal sequences, or erroneous arithmetic aggregations, even when the source text explicitly contains the requisite information. In high-stakes domains, 
such as finance~\citep{kang2023deficiency}, law~\citep{dahl2024large}, and healthcare~\citep{kim2025medical}, 
hallucinations are more prohibitive, as users require LLMs to function as meticulous evidence readers rather than creative generators of plausible but unverified narratives.


Among approaches in mitigating hallucinations, Supervised Fine-Tune (SFT) has been primarily considered as effective~\citep{elaraby2023haloestimationreductionhallucinations,lin2022truthfulqameasuringmodelsmimic,zhou2023limaalignment}. However, several practices emphasize that SFT could also amplify hallucinations by fundamentally operating on behavior cloning~\citep{torabi2018behavioral}, which induces models to prioritize authoritative stylistic patterns over factual accuracy~\citep{casper2023openproblemsfundamentallimitations,zhang2025sirenssongaiocean,gekhman2024doesfinetuningllmsnew}.
Consequently, Reinforcement Learning (RL) has become a necessity for optimizing LLMs towards more truthfulness
~\citep{yang2025qwen3technicalreport,gemmateam2025gemma3technicalreport}. 
However, the standard outcome-level RL rewards lack the granularity required to supervise fine-grained factual consistency, leading to a fundamental misalignment between the scalar reward and the nuanced evidentiary requirements of RAG.
Specifically, Reinforcement Learning from Human Feedback (RLHF) \citep{ouyang2022training} typically relies on reward models (RMs) \citep{bradley1952rank,li2025eliminatinginductivebiasreward,li-etal-2025-aplot,liu2024skyworkrewardbagtricksreward} that compress multifaceted evaluations into coarse-grained scalar scores. This loss of granularity limits the ability to verify individual claims, providing insufficient signals to reinforce the meticulous evidentiary grounding required for data-intensive tasks. Alternatively, while Reinforcement Learning with Verifiable Reward (RLVR) \citep{deepseekai2025deepseekr1incentivizingreasoningcapability} attempts to provide objective feedback via deterministic rules \citep{shao2024deepseekmathpushinglimitsmathematical}, rubrics \citep{gunjal2025rubricsrewardsreinforcementlearning}, or external generalist reward model~\citep{liu2025inferencetimescalinggeneralistreward}, it remains fundamentally constrained by a dual-dependency bottleneck: the scarcity of expert-annotated groundtruth in specialized domains, and the reasoning ceilings of external verifiers. Consequently, unsupported intermediate claims can persist, as models are rewarded for correct final outcomes despite potentially hallucinated reasoning trajectories~\citep{li2025reasoningmodelshallucinatemore}.


While post-hoc metrics and fact-checkers \citep{tang2024minicheck,rashkin2023measuring,gao-etal-2023-rarr,baker2025monitoringreasoningmodelsmisbehavior,guan2025monitoringmonitorability} provide utility for evaluation, such static tools are seldom integrated into the internal optimization loop to provide actionable, claim-oriented guidance. More critically, the efficacy of such verifiers is fundamentally undermined by the 
\textit{information leakage} inherent in current evaluation schemes \citep{seo2025verifyingverifiersunveilingpitfalls}. 
Specifically, concurrent exposure to the query, documents, and generated response induces a confirmation bias~\citep{sharma2025understandingsycophancylanguagemodels}. In this state, verifiers tend to prioritize validating internal coherence over objective grounding against source evidence, leading to the unwarranted endorsement of erroneous claims. This compromised ``blind scrutiny'' prevents the model from internalizing the rigorous, document-anchored reasoning essential for high-fidelity generation.



To address above factuality challenges, we introduce \textbf{M}ulti-\textbf{A}gent \textbf{R}einforced self-\textbf{C}heck for \textbf{H}allucination (MARCH), an RL-based framework that enforces factual alignment through deliberate information asymmetry. 
MARCH orchestrates a collaborative pipeline of three specialized agents: a \textit{Solver}, a \textit{Proposer}, and a \textit{Checker}, to prevent model-based verifiers from taking cognitive shortcuts through confirmation bias.
%
%
As illustrated in Figure~\ref{fig:framework_1}, the Solver first generates a response based on the query and retrieved documents. Subsequently, the Proposer acts as a 'Response Atomizer,' deconstructing the narrative into discrete, verifiable claims. Finally, the Checker validates these claims by re-answering them exclusively based on the source documents, strictly blinded to the Solver’s original output to avoid cognitive contamination. Through this asymmetric decoupling, MARCH transforms the verification process from a circular confirmation of plausible text into a rigorous cross-examination between generated claims and factual evidence.
%
%
During training, MARCH promotes agentic self-evolution via a zero-tolerance reward mechanism: any discrepancy between the Proposer’s extracted claims and the Checker’s document-based validation triggers a penalty on the entire response trajectory. This encourages the agents to align their outputs strictly with retrieved evidence, effectively internalizing the capability for rigorous, document-anchored statements.
%
Experimentally, MARCH achieves a substantial reduction in hallucinations compared to its base model, without requiring additional human annotations or external fact-checking tools, thereby demonstrating the efficacy of our asymmetric collaborative reinforcement paradigm. Notably, an 8B-parameter LLM equipped with MARCH achieves performance parity with leading closed-source models on multiple hallucination benchmarks, while simultaneously delivering significant gains in general RAG-QA tasks. These results establish a scalable and verifiable pathway toward trustworthy, agentic self-improvement for large language models.

\section{Methodology}
\subsection{Problem Formulation and Overview}
Given an input query $\vx \in \gX$ and a set of retrieved documents $\mD = [\vd_1, \vd_2, \dots, \vd_l]$, we aim to optimize a policy model $\pi_{\theta}(\vy|\vx, \mD)$ that generates a response $\vy$ with maximum factual grounding aligned with the provided evidences in $\mD$. Formally, suppose we can extract all the statements of facts $\{\va_1,\va_2,\dots,\va_n\}$ from the response $\vy$ using a function $\gA(\cdot)$, as $\gA(\vy) = \{\va_i\}_{i=1}^n$. Then, from the perspective of reinforcement learning with verifiable reward (RLVR), if an oracle set of ground-truth annotations $\mathcal{A}^* = \{\va^*_1, \dots, \va^*_n\}$ is ideally available for every atomic statement $\va_i$ in $\vy$, the optimization objective is to maximize the expected evidentiary consistency:
\begin{equation}\label{eq:oracle_objective}
   \max_{\theta} \mathbb{E}_{\vx\sim \gX, \vy \sim \pi_{\theta}(\cdot | \vx, \mD)} [ R(\{\va_i, \va^*_i\}_{i=1}^n) ],
\end{equation}
where $R(\cdot)$ is a fact-check function that identifies the exact match between $\mathcal{A}(\vy) = \{\va_i\}_{i=1}^n$ and $\{\va^*_i\}_{i=1}^n$. However, $\mathcal{A}^*$ is practically unavailable in most application domains and requires enormous human annotation. 

Therefore, we reformulate \eqref{eq:oracle_objective} into a self-contained, verifiable objective by decomposing the grounding process into three separate agent tasks, as illustrated in Figure~\ref{fig:framework_1}: 
\begin{enumerate}[leftmargin=2.5ex]
\vspace{-1mm}
    \item \textbf{Response Generation}: a Solver agent ordinarily responses $\vy$ based on the input query $\vx$ and retrieved context $\mD$. 
    \item  \textbf{Claim-level QA Proposal}: a Proposer agent decomposes the response $\vy$ into $n$ discrete factual claims in a question-answering (QA) format, as $\gQ(\vy) = \{(\vq_i, \va_i)\}_{i=1}^n$.
    \item \textbf{Factual Verification}: a Checker agent answers all the atomic questions $\{\vq_i\}_{i=1}^n$ as $\{\hat{\va}_i\}_{i=1}^n$ with only the retrieval documents $\mD$ provided, without any access to the original response $\vy$ and extracted answers$\{\va_i\}_{i=1}^n$. 
\end{enumerate}
By substituting the unreachable oracle $\va^*_i$ with the  grounding result $\hat{\va}_i$ from the multi-agent pipeline, we approximate the factuality optimization as: 
\begin{equation}\label{eq:integrated_objective_vanilla}
   \max_{\theta} \mathbb{E}_{\vx\sim \gX, \vy \sim \pi_{\theta}(\cdot | \vx, \mD)} \left[ R(\{(\va_i, \hat{\va}_i)\}_{i=1}^n) \right].
\end{equation}
By optimizing \eqref{eq:integrated_objective_vanilla}, we force $\pi_\theta$ to internalize a rigorous verification logic, ensuring that the generated content is strictly anchored in the source evidence to enhance overall factual accuracy. We call the proposed learning scheme \textbf{M}ulti-\textbf{A}gent \textbf{R}einforced self-\textbf{C}heck for \textbf{H}allucination (MARCH).

\subsection{MARCH Modeling}\label{met:modeling}

As shown in Figure~\ref{fig:framework_1}, we instantiate the functional primitives identified above into three specialized agents derived from the same base policy $\pi_\theta$: a \textit{Solver}, a \textit{Proposer}, and a \textit{Checker}, which are conditioned to exhibit distinct epistemic behaviors, effectively transforming the RAG task into a structured, asymmetric cooperative game. MARCH begins with the {Solver}'s policy:
\begin{equation}
   \nu_{\text{solve}}(\cdot | \vx, \mD)  =  \pi_{\theta}(\cdot | \vx, \mD,  \vs_\text{solve}),
\end{equation} where $\vs_\text{solve}$ is the Solver's system prompt (as shown in Appendix~\ref{app:answer_prompt}) to induce the policy model $\pi_\theta$ to generate a comprehensive response $\vy$ given the retrieval documents $\mD$ and the input $\vx$.

To decouple the continuous response $\vy$ into verifiable components, the {Proposer} agent functions as a \textit{Response Atomizer} that bridges narrative synthesis and discrete verification:
\begin{equation}
  \nu_{\text{propose}}(\cdot | \vy) = \pi_{\theta}(\cdot|  \vy, \vs_\text{propose}),
\end{equation}
where the Propser's system prompt $\vs_\text{propose}$ is in Appendix~\ref{app:propose_qa}.
Specifically,  $\nu_{\text{propose}}$ decomposes the response $\vy$ into a set of $n$ self-contained factual question-answer (QA) pairs $\mathcal{Q}(\vtau) = \{(\vq_i, \va_i)\}_{i=1}^n$, where each $\vq_i$ represents a targeted inquiry,  and each $\va_i$ denotes the corresponding claim asserted in $\vy$. The function $\gQ(\cdot)$ extracts the QA pairs from the Proposer's generation $\vtau \sim \nu_\text{propose}(\cdot|\vy)$.  
the {Checker} operates as the \textit{Blinded Auditor} and the linchpin of the asymmetric cooperation:
\begin{equation}
    \nu_{\text{check}}(\cdot |\{\vq_i\}_{i=1}^n, \mD) = \pi_{\theta}(\cdot| \{ \vq_i\}, \mD, \vs_\text{check})
\end{equation}
with the Checker's prompt $\vs_\text{check}$ in Appendix~\ref{app:check_qa}
To eliminate confirmation bias and break information symmetry, when tasked with answering $\vq_i$,  $\nu_{\text{check}}$ is strictly {blinded} to the Solver's response $\vy$ and the corresponding claims $\va_i$. 
the \textit{Checker} replies all the queries $\{ \vq_i\}_{i=1}^n$ in a response $\bm{\lambda}\sim \nu_\text{check}(\cdot|\{\vq_i\}_{i=1}^n, \mD)$, only based on the RAG documents $\mD$. The corresponding answers $\gA(\bm{\lambda}) = \{\hat{\va}_i\}_{i=1}^n$ are parsed by the function $\gA(\cdot)$ to serve as the re-checked answers to the QA pairs $\{(\vq_i, \va_i)\}_{i=1}^n$.

\paragraph{Overall Objective.} By integrating the conditional dependencies of the \textit{Solver},  the \textit{Proposer}, and the \textit{Checker}, MARCH defines a unified {asymmetric cooperative and play objective} by focusing on the {evidentiary identity} between \textit{Solver's} response $\vy$ and \textit{Checker's} independent grounding trajectory $\bm{\lambda}$, mediated by the atomic claim decomposition of the \textit{Proposer}. The optimization goal is to maximize the expected reward based on the consistency of claims, marginalized over the discrete atomization performed by $\nu_\text{propose}$:
\begin{align}
   &\max_{\pi_\theta}\ \mathbb{E}_{
   \vx\sim\gX, \vy \sim \nu_{\text{solve}} (\cdot|\vx, \mD)} \left[ R(\{(\va_i, \hat{\va}_i)\}_{i=1}^n) \right] \label{eq:integrated_objective} \\
\text{s.t. }
   \{(\vq_i, \va_i)\}_{i=1}^n = \gQ(\vtau)&, \ \vtau\sim \nu_{\text{propose}}(\cdot|\vy) \{\hat{\va}_i\}_{i=1}^n = \gA(\bm{\lambda}),   \ \bm{\lambda} \sim \nu_{\text{check}}(\cdot|\{\vq_i\}_{i=1}^n, \mD)  \nonumber
\end{align}   
%
where $R(\cdot)$ measures the evidentiary identity between each asserted claim $\va_i$ and the independent grounding $\hat{\va}_i$. By optimizing \eqref{eq:integrated_objective}, the policy $\pi_\theta$ is compelled to internalize a rigorous verification logic, ensuring that its narrative synthesis is strictly anchored in verifiable evidence to enhance overall factual accuracy, which also develops a self-consistent grounding capability that eliminates reliance on external supervision.

\subsection{MARCH Implementation}\label{met:implementation}
In our implementation, we specifically prioritize {numerical and quantitative verification}, driven by the observation that in data-intensive RAG scenarios, LLMs frequently suffer from hallucinations such as misquoting figures or altering numerical values during context integration. By anchoring the verification loop on these high-fidelity tokens, MARCH provides a more rigorous evidentiary baseline for grounding long-form reasoning.
\paragraph{Zero-Tolerance Reward (ZTR).}
To enforce absolute factual grounding, we define a binary {Zero-Tolerance Reward} function $R(\{(\va_i, \hat{\va}_i)\}_{i=1}^n)$ in \eqref{eq:integrated_objective}. In the RAG context, we posit that any single hallucinated claim $\va_i$ within the response $\vy$ invalidates the entire reasoning chain. Therefore, a trajectory is assigned a successful score only if every claim $\va_i$ asserted by the \textit{Solver} achieves an identity match with the consensus answer $\hat{\va}_i$ extracted by the \textit{Checker}:
\begin{equation}\label{eq:ztr_reward}
    R(\{(\va_i, \hat{\va}_i)\}_{i=1}^n) = 
    \begin{cases} 
    0, \hspace{-1.5ex}&\text{if every } \va_i = \hat{\va}_i 
    \\
    -1, \hspace{-1.5ex}&\text{otherwise}
    \end{cases}
\end{equation}
%
%
This ``all-or-nothing'' reward structure prevents the policy $\pi_\theta$ from optimizing for partial correctness or stylistic plausibility, forcing the model to prioritize strict evidentiary identity. Besides, to ensure the \textit{Checker} provides a reliable baseline for the audit and to reduce the variance of the reward signal, we employ a multi-sampling strategy during the cross-validation phase. For each question $\vq_i$ generated by the \textit{Proposer},  $\nu_\text{check}$ generates multiple independent audit trajectories. We apply a majority voting to determine the final consensus answer $\hat{\va}_i$, which effectively mitigates stochastic errors in \textit{Checker's} generation in one independent audit.
\paragraph{Policy Optimization.}
We optimize the shared policy $\pi_\theta$ of the three agents using Proximal Policy Optimization (PPO) \citep{schulman2017proximalpolicyoptimizationalgorithms}. 
Critically, we incorporate both the reasoning path $\vy$ and the audit trajectory $\bm{\lambda} \sim \nu_\text{check}(\cdot|\{\vq_i\}, \mD)$ into the optimization loop, ensuring that $\pi_\theta$ concurrently evolves as both a reliable generator and a rigorous auditor.
Consequently, each original query of a training batch $\vx_i \in \mathcal{B}$ contributes two distinct reward signals to the policy, allowing $\pi_\theta$ to learn the fine-grained correspondence between narrative synthesis and independent document-grounding within a unified training batch. While both trajectories derive their terminal reward from the same evidentiary identity match (\eqref{eq:ztr_reward}), corresponding advantages $\hat{A}$ and KL-divergence~\citep{csiszar1975divergence} penalties against the reference model $\pi_\text{ref}$ are estimated independently. The joint gradient is computed as the sum of gradients across all individual trajectories $\nabla_\theta \mathcal{L}_{\text{MARCH}}(\theta) \approx$
\begin{align}\label{eq:expanded_gradient}
    \frac{1}{2|\gB|} \sum_{\vx \in \mathcal{B}} \Bigg[\sum_{t=1}^{|\vy|} \hat{A}_{y}^t  \nabla_\theta \log \pi_\theta({y}^t | \cdot)  +\sum_{t=1}^{|\bm{\lambda}|} \hat{A}_{\lambda}^t   \nabla_\theta \log \pi_\theta({\lambda}^t | *)
    - \beta \nabla_\theta \text{KL}[\pi_\theta \| \pi_{\text{ref}}] \Bigg]
\end{align}
where $\hat{A}_\lambda^t$  and  $\hat{A}_y^t$ are the $t$-th token's corresponding advantages of the Checker's output $\bm{\lambda}$ and the Solver's output $\vy$, estimated via Generalized Advantage Estimation~\citep{schulman2015high}. The notation $(\cdot)$ and $(*)$ in \eqref{eq:expanded_gradient} encapsulates the respective contexts for brevity: $(\vx, \mD, \vy^{<t})$ and $(\{\vq_i\}, \mD, \bm{\lambda}^{<t})$ for the Solver's and the Checker's generation prefixes. \eqref{eq:expanded_gradient} ensures that $\pi_\theta$ concurrently refines its narrative generation and evidentiary precision, internalizing a self-contained verification loop where generative assertions are inherently tethered to the policy's own auditing capacity. We summarize our training algorithm in Algorithm~\ref{alg:training}.

\begin{algorithm*}[!t]
\textbf{Input}: An LLM Policy $\pi_\theta$; A Training Batch $\mathcal{B}=\{(\vx_i,\mD_i)\}_{i=1}^{|\mathcal{B}|}$; Solver, Proposer and Checker System Prompts $\{\vs_{\text{solve}}, \vs_{\text{propose}}, \vs_{\text{check}}\}$\;
\tcp{Phase 1: Multi-Agent Collaboration and Rollout}
\For{each training sample $(\vx, \mD) \in \mathcal{B}$}{
    \tcp{Step 1: Generate reasoning path $\vy$ from Solver}
    Sample $\vy \sim \nu_{\text{solve}}(\cdot\mid\vx,\mD) = \pi_{\theta}(\cdot | \vx, \mD, \vs_{\text{solve}})$\;
    
    \tcp{Step 2: Atomize response $\vy$ by Proposer and Parse QA pairs by $\mathcal{Q}(\cdot)$}
    Sample $\boldsymbol{\tau}\sim\nu_{\text{propose}}(\cdot\mid\vy) = \pi_{\theta}(\cdot \mid \vy, \vs_{\text{propose}})$\;
    Extract QA pairs $\{(\vq_i,\va_i)\}_{i=1}^{n}=\mathcal{Q}(\boldsymbol{\tau})$\;
    
    \tcp{Step 3: Generate audit trajectory $\bm{\lambda}$ from Checker and Parse re-checked answers by $\mathcal{A}(\cdot)$}
    Sample $\bm{\lambda}\sim\nu_{\text{check}}(\cdot\mid\{\vq_i\}_{i=1}^n,\mD) = \pi_{\theta}(\cdot \mid \{\vq_i\}_{i=1}^n,\mD, \vs_{\text{check}})$\;
    Parse re-checked answers $\{(\hat{\va}_i)\}_{i=1}^{n}=\mathcal{A}(\bm{\lambda})$\;
    
    \tcp{Step 4: Unified reward labeling}
    Calculate Zero-Tolerance Reward $R_i$ by \eqref{eq:ztr_reward} (based on the match between $\va$ and $\hat{\va}$)\;
    
    \tcp{Step 5: Store both reasoning and auditing trajectories}
    Add $\vy_i$ and $\bm{\lambda}_i$ to the unified batch trajectory set $\mathcal{T}_{\mathcal{B}}$\;
    Associate both $\vy_i$ and $\bm{\lambda}_i$ with terminal reward $R_i$\;
}

\tcp{Phase 2: Joint Policy Optimization}
\For{each trajectory $\tau' \in \mathcal{T}_{\mathcal{B}} = \{\vy_i, \bm{\lambda}_i\}$}{
    Estimate Advantage $\hat{A}_{\tau'}$ using GAE based on $R_i$\;
    Compute ${\text{KL}}(\pi_\theta \| \pi_{\text{ref}})$ for each token in $\tau'$\;
}
Update $\pi_\theta$ by computing the joint gradient $\nabla_\theta \mathcal{L}_{\text{MARCH}}(\theta)$ according to \eqref{eq:expanded_gradient}\:
$    \frac{1}{2|\gB|} \sum_{\vx \in \mathcal{B}} \Bigg[\sum_{t=1}^{|\vy|} \hat{A}_{y}^t  \nabla_\theta \log \pi_\theta({y}^t | \cdot) +\sum_{t=1}^{|\bm{\lambda}|} \hat{A}_{\lambda}^t   \nabla_\theta \log \pi_\theta({\lambda}^t | *) - \beta \nabla_\theta \text{KL}[\pi_\theta \| \pi_{\text{ref}}] \Bigg]$\;
where the notation $(\cdot)$ and $(*)$ in \eqref{eq:expanded_gradient} encapsulates the respective contexts for brevity: $(\vx, \mD, \vy^{<t})$ and $(\{\vq_i\}, \mD, \bm{\lambda}^{<t})$ for the Solver's and the Checker's generation prefixes\;
\caption{MARCH Training Process.}
\label{alg:training}
\end{algorithm*}


\section{Experiment}
We first evaluate the effectiveness of MARCH in mitigating hallucinations.
Then, we show that the improved factual consistency ability helps downstream reasoning tasks like Question-Answering.
\subsection{Training Setup}
\paragraph{Training Datasets and Base Model.}
We train our framework covering: a {STEM} (Science-Technology-Engineering-Math) setting based on BioASQ~\citep{krithara2023bioasq}, and a {General} setting based on 2WikiMultiHopQA~\citep{ho2020constructing} and MuSiQue~\citep{trivedi2022musique}. Crucially, training is performed using only queries and retrieved documents, without access to ground-truth answers or annotations. Table~\ref{tab:training_datasets_stats} shows the number of retrieved documents per query and the document lengths vary widely across datasets, and a substantial fraction of the documents are irrelevant to the query (e.g., 30.1\% for STEM and up to 88.4\% for MuSiQue). Such a high level of noise and heterogeneity in the retrieval docs makes the training setting highly realistic but also extremely challenging, as the agents must learn to perform document-grounded fact checking in the presence of many noisy documents. We consider Meta-Llama3.1-8B-Instruct~\citep{touvron2023llamaopenefficientfoundation} as the initialization policy.

\begin{table*}[!h]
\centering
\caption{Statistics of the training dataset, where we report the number of training samples, average query length, the number of retrieved documents per query, the average document length, and the proportion of documents that are irrelevant to the query. The large variance in document counts and lengths, together with the high ratio of irrelevant documents, highlights the realism and difficulty of our training setup.}
\resizebox{\textwidth}{!}{
\begin{tabular}{ccccccccccc}
\toprule
\multirow{2}{*}{Dataset} & \multirow{2}{*}{Original Dataset} & \multirow{2}{*}{\#Samples} & \multirow{2}{*}{Avg. Query Len} & \multicolumn{3}{c}{\#Docs per Query} & \multicolumn{3}{c}{Total Doc Len (words)} & \multirow{2}{*}{Irrelevant Doc Ratio} \\ 
 &  &  &  & Avg & Max & Min & Avg & Max & Min &  \\ \midrule
STEM & BioASQ & 4721 & 8.62 & 9.03 & 157 & 1 & 1314.86 & 22135 & 1 & 30.14\% \\ \midrule
\multirow{2}{*}{General} & 2WikiMultiHopQA & 4500 & 12.87 & 10 & 10 & 10 & 575.65 & 2099 & 95 & 24.4\% \\
 & MuSiQue & 4500 & 15.85 & 19.99 & 20 & 16 & 11221.72 & 20741 & 4112 & 88.35\% \\ \bottomrule
\end{tabular}
}\label{tab:training_datasets_stats}
\end{table*}

\paragraph{Training Details.}
Training framework is built upon the \texttt{VerL}~\citep{Sheng_2025}, and we employ Fully Sharded Data Parallel (FSDP)~\citep{zhao2023pytorchfsdpexperiencesscaling} for efficient multi-node multi-GPU training. During the training rollout, we utilize the \texttt{vLLM}~\citep{kwon2023efficientmemorymanagementlarge} for efficient sample generation with the temperature of 0.6. The PPO~\citep{schulman2017proximalpolicyoptimizationalgorithms} training is conducted for a single epoch with a global batch size of 32, with a max prompt length of 24,567, max response length of 8,192. We set learning rates for the actor and the critic as $1\times10^{-6}$ and $1\times10^{-5}$, respectively. We save and evaluate a checkpoint every 20 steps, and report results from the single best-performing checkpoint selected based on the highest average score across all evaluation benchmarks. More detailed training hyper-parameters are in Appendix~\ref{app:training_hp}.



\subsection{Experiments on Hallucination}
\paragraph{Evaluation Benchmark.}
We evaluate MARCH along the aspect of \emph{hallucination and factual grounding} on the RAGTruth~\citep{niu2024ragtruth}, FaithBench~\citep{tamber2025benchmarking}, ContextualJudgeBench~\citep{xu2025does} and Facts Grounding benchmark~\citep{jacovi2025facts}, which cover QA, data-to-text writing, summarization, and diverse domains such as medical, financial, and legal. Detailed introductions are provided in Appendix~\ref{app:eval-details-fc}. The best performance is marked with \rr{Red} and the second-best with \bb{Blue}.

\paragraph{Evaluation Protocol and Metrics.}
Following~\citet{niu2024ragtruth,tamber2025benchmarking,jacovi2025facts,xu2025does}, we focus on \emph{response-level} factuality eval. Given generation, retrieved documents, and ground-truth reference, we deploy a Qwen3-235B-A22B~\citep{yang2025qwen3technicalreport} as a judge model to decide whether the response is factually consistent/accurate with the evidence/ground-truth factual labels. We mitigate the randomness during evaluation by drawing {eight} independent generations for each test query; the judge model then evaluates each sampling individually, and the final verdict is determined by a majority vote across these eight trajectories.

\paragraph{Results on RAGTruth and FaithBench.} Table~\ref{tab:res_main_ragtruth} shows that our method brings substantial and consistent improvements over the base Llama3.1-8B-Instruct across {all} settings. With STEM data (\textit{MARCH-STEM}), the learned policy improves the average accuracy from 55.20 to 74.93 ({+19.73}) and with General QA data (\textit{MARCH-General}), the average accuracy further rises to 75.23 ({+20.03}), which suggest that MARCH reliably strengthens the base model’s ability to judge and mitigate hallucinations, rather than overfitting to a particular domain. When compared to other larger or proprietary LLMs, our policies also achieve an advanced average performance. Together, these results demonstrate that our framework can turn a small-sized base LLM into a highly competitive factuality judge simply by training on noisy, document-grounded RAG data.

\begin{figure}[!h]\vspace{-1em}
    \centering
    \includegraphics[width=.7\linewidth]{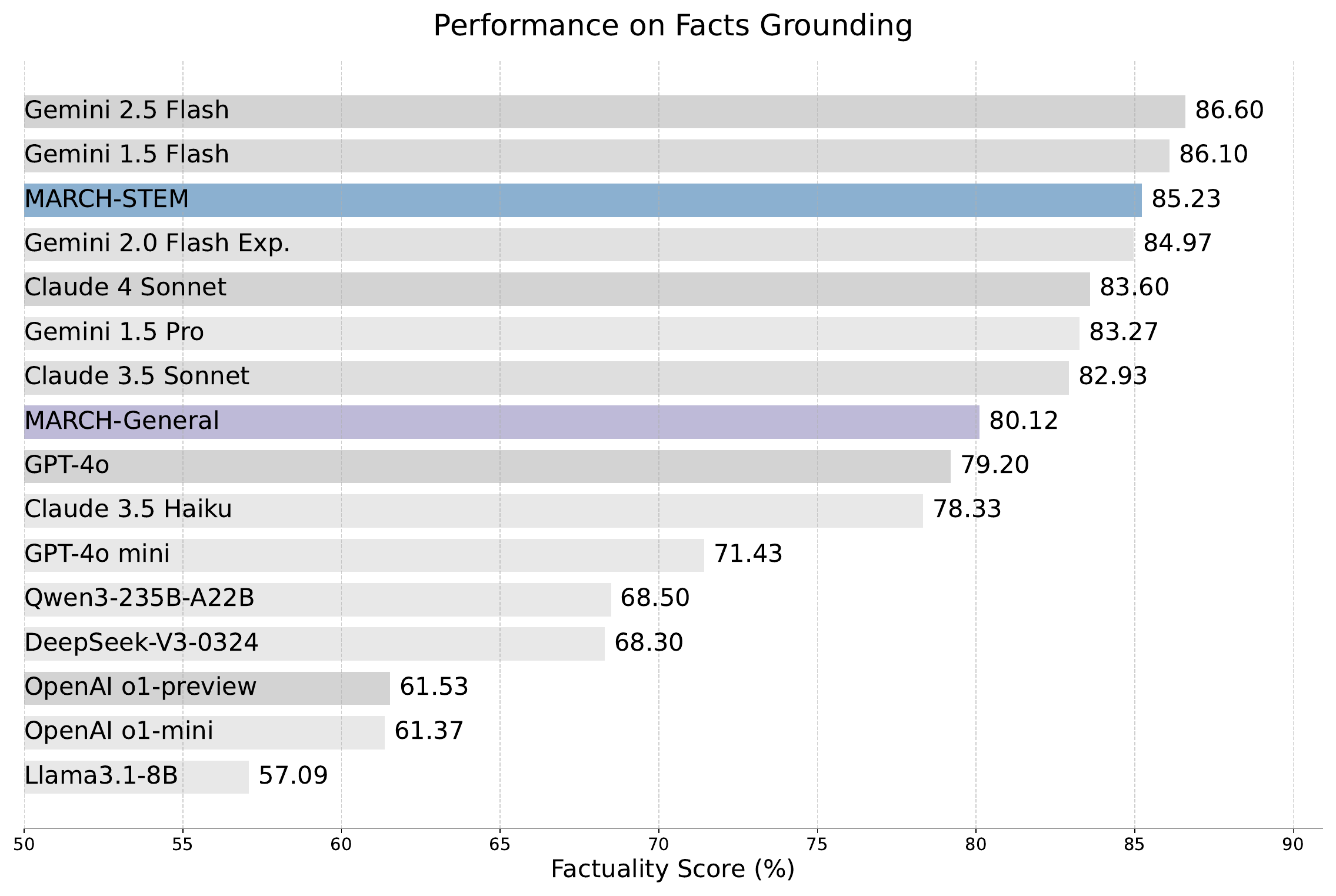}
    \caption{Comparison of performance on the Facts Grounding. MARCH (highlighted in blue and purple) demonstrates competitive performance against a range of leading open and proprietary models.}
    \label{fig:main_res_fg}
\end{figure}

\begin{table}[!h]
\centering
\caption{Consistency Rate (\%) on RAGTruth and FaithBench benchmark. Baseline results are cited from~\url{https://github.com/vectara/FaithJudge/tree/main/eval_results}.}
\resizebox{0.7\textwidth}{!}{
\begin{tabular}{l|ccccc}
\toprule
\multirow{2}{*}{Model} & \multicolumn{3}{c}{RAGTruth} & \multirow{2}{*}{FaithBench}  & \multirow{2}{*}{Average} \\
 & Summary & Data2Txt & QA &  &  \\ \midrule
o4-mini-high-2025-04-16 & 76.67 & 64.00 & 82.01&  45.83 & 67.13 \\
o4-mini-medium-2025-04-16 & 76.67 & 67.33 & 88.49& 55.56  & {71.01} \\
o4-mini-low-2025-04-16 & 70.00 & 71.33 & 83.45& 55.56  & 70.09 \\
GLM-4-9B-Chat & \bb{87.77} & 52.67 & 90.67& \rr{61.11}  & 73.06 \\
Phi-4-mini-Instruct & 78.67 & 24.67 & \rr{94.96}& 38.89 & 59.30 \\
Qwen2.5-14B-Instruct & 82.00 & 51.33 & \bb{93.53}& 45.83  & 68.17 \\
Ministral-8B-Instruct & 83.33 & 49.33 & 88.49& 43.06 & 66.05 \\ \midrule
Meta-Llama3.1-8B-Instruct & 71.33 & 48.67 & 63.31& 37.50 & 55.20 \\
\rowcolor{blue!5}\textbf{MARCH-STEM} & 86.67 & \rr{72.67} & 83.45& \bb{56.94}  & \bb{74.93} \\
\rowcolor{blue!5}$\Delta$ & \textcolor{ggg}{$\uparrow15.34$} & \textcolor{ggg}{$\uparrow24.00$} & \textcolor{ggg}{$\uparrow20.14$} & \textcolor{ggg}{$\uparrow19.44$} & \textcolor{ggg}{$\uparrow19.73$}\\
\rowcolor{blue!5}\textbf{MARCH-General} & \rr{92.67} & \bb{72.00} & 83.45 & 52.78 & \rr{75.23} \\
\rowcolor{blue!5}$\Delta$ & \textcolor{ggg}{$\uparrow21.34$} & \textcolor{ggg}{$\uparrow23.33$} & \textcolor{ggg}{$\uparrow20.14$} & \textcolor{ggg}{$\uparrow15.28$} & \textcolor{ggg}{$\uparrow20.03$} \\ \bottomrule
\end{tabular}
}\label{tab:res_main_ragtruth}
\end{table}
\paragraph{Results on Facts Grounding.} FACTS Grounding evaluates whether model responses are fully supported by a single source document. Figure~\ref{fig:main_res_fg} shows that our policies achieve large gains over the base model, reaching a factuality score of {85.23\%} (MARCH-STEM) and {80.12\%} (MARCH-General), compared with {57.09\%} for the base Llama3.1‑8B-Instruct. Our method successfully lifts an 8B backbone into the comparable performance band as leading proprietary models, placing \textit{MARCH‑STEM} just behind the top Gemini 2.5/1.5 Flash models on the public leaderboard while surpassing popular models such as GPT‑4o and several larger open models. The strong results indicate that MARCH substantially improves long‑document grounding robustness across heterogeneous domains. 
\begin{table*}[!h]
\centering
\caption{Accuracy (\%) on ContextualJudgeBench. Baseline results are cited from~\citet{xu2025does}.}
\resizebox{\textwidth}{!}{
\begin{tabular}{l|ccccccccc}
\toprule
 \multirow{2}{*}{Model}  & {Refusal} & {Refusal} & {Faithfulness} & {Faithfulness} & {Completeness} & {Completeness} & {Conciseness} & {Conciseness} & \multirow{2}{*}{{Avg.}} \\
 & {(Ans.)} & {(Unans.)} & {(QA)} & {(Summ.)} & {(QA)} & {(Summ.)} & {(QA)} & {(Summ.)} & \\
\midrule
\rowcolor{mygray} \multicolumn{10}{c}{\textit{Small Judge Model}} \\
Glider-3.8B~\citep{deshpande2024glidergradingllminteractions} & 12.0 & 8.8 & 45.6 & 9.2 & 20.8 & 28.7 & 5.1 & 4.1 & 16.8 \\
Promtheus-2-7b~\citep{kim-etal-2024-prometheus} & 12.4 & 44.0 & 27.2 & 32.0 & 24.0 & 42.6 & 6.7 & 29.5 & 27.3 \\
Llama-3-OffsetBias-8B~\citep{park-etal-2024-offsetbias} & 64.8 & 11.2 & 34.0 & 26.4 & 33.2 & 21.1 & 46.3 & 23.0 & 32.6 \\
Skywork-8B~\citep{liu2024skyworkrewardbagtricksreward} & 60.8 & 12.0 & 38.8 & 31.6 & 38.4 & 26.7 & 29.4 & 21.3 & 32.4 \\
Alta-Selene-8B~\citep{alexandru2025atlaseleneminigeneral} & 74.4 & 26.4 & 40.8 & 32.8 & 32.4 & 34.7 & 23.1 & 32.0 & 37.1 \\
SFRJudge-8B~\citep{wang2025directjudgementpreferenceoptimization} & 70.8 & 22.0 & 40.4 & 38.8 & 40.4 & 43.4 & 27.5 & 31.1 & {39.3} \\
SFRJudge-12B~\citep{wang2025directjudgementpreferenceoptimization} & 68.4 & 28.4 & 45.2 & 43.6 & 28.0 & 51.0 & 16.1 & 29.5 & 38.8 \\
\midrule
\rowcolor{mygray} \multicolumn{10}{c}{\textit{ Large Judge Model}} \\
Promtheus-2-8x7b~\citep{kim-etal-2024-prometheus} & 22.0 & 29.6 & 22.4 & 29.6 & 20.4 & 39.8 & 10.2 & 18.4 & 24.1 \\
Skywork-70B~\citep{liu2024skyworkrewardbagtricksreward} & 82.4 & 11.2 & 48.0 & 47.6 & 36.8 & 41.4 & 21.6 & 27.9 & 39.6 \\
ST-Eval-70B~\cite{wang2024selftaughtevaluators} & 50.0 & 42.0 & 51.2 & 45.6 & 40.8 & 39.4 & 36.1 & 29.9 & 41.9 \\
SFRJudge-70B~\citep{wang2025directjudgementpreferenceoptimization} & 87.6 & 32.4 & 60.8 & \bb{54.8} & 40.8 & 53.4 & \bb{44.7} & 36.1 & {51.4} \\
\midrule
\rowcolor{mygray} \multicolumn{10}{c}{\textit{Instruct + Reasoning Model}}   \\
Llama-3.1-70B~\citep{touvron2023llamaopenefficientfoundation} & 59.6 & 48.0 & 58.0 & 48.4 & 38.0 & 51.8 & 15.7 & 27.5 & 43.4 \\
Llama-3.3-70B~\citep{touvron2023llamaopenefficientfoundation} & 71.6 & 42.4 & 68.0 & 48.4 & 42.0 & 51.8 & 20.8 & 30.7 & 47.0 \\
R1Distill-Llama-3.3-70B~\citep{deepseekai2025deepseekr1incentivizingreasoningcapability} & \bb{89.6} & 50.4 & \rr{74.0} & 48.4 & 42.4 & 57.4 & 19.2 & 29.5 & 51.4 \\
GPT-4o-mini & 71.2 & 22.8 & 45.6 & 42.4 & 33.2 & 54.2 & 11.8 & 29.5 & 38.8 \\
GPT-4o & 64.0 & 52.0 & 68.0 & 50.8 & 39.6 & 56.2 & 12.9 & 22.5 & 45.8 \\
DeepSeek-R1~\citep{deepseekai2025deepseekr1incentivizingreasoningcapability} & \rr{92.0} & 52.0 & \bb{72.0} & 50.4 & 41.2 & 60.6 & 20.4 & 26.2 & \bb{51.9} \\
\midrule
\rowcolor{mygray} \multicolumn{10}{c}{\textit{OURS}}   \\
Llama-3.1-8B & 28.0 & 43.2 & 34.8 & 34.8 & 23.2 & 41.0 & 11.4 & 21.3 & 29.7 \\
\rowcolor{blue!5}\textbf{MARCH-General}       & 37.2 & 57.8 & 52.6 & \bb{54.8} & \rr{48.0} & \bb{63.9} & \rr{48.1} & \rr{50.1} & {51.6} \\
\rowcolor{blue!5}$\Delta$ & \textcolor{ggg}{$\uparrow9.2$} & \textcolor{ggg}{$\uparrow14.6$} & \textcolor{ggg}{$\uparrow17.8$} & \textcolor{ggg}{$\uparrow20.0$} & \textcolor{ggg}{$\uparrow24.8$} & \textcolor{ggg}{$\uparrow22.9$} & \textcolor{ggg}{$\uparrow36.7$} & \textcolor{ggg}{$\uparrow28.8$} & \textcolor{ggg}{$\uparrow21.9$} \\ 
\rowcolor{blue!5}\textbf{MARCH-STEM}      & 38.1 & \rr{64.0} & 53.6 & \rr{58.0} & 44.9 & \rr{65.5} & \bb{44.7} & \bb{49.5} & \rr{52.3} \\
\rowcolor{blue!5}$\Delta$ & \textcolor{ggg}{$\uparrow10.1$} & \textcolor{ggg}{$\uparrow20.8$} & \textcolor{ggg}{$\uparrow18.8$} & \textcolor{ggg}{$\uparrow23.2$} & \textcolor{ggg}{$\uparrow21.7$} & \textcolor{ggg}{$\uparrow24.5$} & \textcolor{ggg}{$\uparrow33.3$} & \textcolor{ggg}{$\uparrow28.2$} & \textcolor{ggg}{$\uparrow22.6$} \\ 
\bottomrule
\end{tabular}
} \label{tab:comp_main_res_cjb}
\end{table*}
\paragraph{Results on ContextualJudgeBench.} Table~\ref{tab:comp_main_res_cjb} reports results on ContextualJudgeBench, which evaluates multiple dimensions of judge behavior, including refusal, faithfulness, completeness, and conciseness. Starting from Llama3.1-8B, our framework again yields large gains across all eight dimensions under both training regimes. Specifically, the policy trained on General QA data (\textit{MARCH-General}) improves the average accuracy from 29.7 to 51.6 ({+21.9}), while the policy trained on STEM data (\textit{MARCH-STEM}) reaches 52.3 ({+22.6}). The improvements are particularly pronounced on faithfulness and completeness (both QA and summarization), where accurate document-grounded judgment is most critical. Based on an 8B LLM, our policies surprisingly reach or surpass the performance of many much larger judge models, and approach the best specialized judge models on several dimensions. Table~\ref{tab:comp_main_res_cjb} indicates that MARCH not only improves binary hallucination detection but also equips LLM with a more nuanced ability to assess answer quality along multiple axes, such as whether key information is covered and whether responses remain concise and well-calibrated in context.

\subsection{Experiments on Multi-Hop QA}
Intuitively, the capacity for precise factual extraction and verification reasonably serves as the fundamental bedrock for complex reasoning, as Question Answering tasks usually require a model to sequentially bridge information across disparate documents. 
In this section, we empirically find that the improved factual grounding brought by MARCH can directly translate into better performance on reasoning-intensive downstream tasks.

\paragraph{Evaluation Benchmarks.} Following~\citet{xu2025collab,trivedi2023interleaving,shao2023enhancing}, we evaluate the performance of our MARCH on three popular multi-hop QA datasets, including HotpotQA~\citep{yang2018hotpotqa}, 2WikiMultiHopQA~\citep{ho2020constructing} and MuSiQue~\citep{trivedi2022musique}. More details about datasets are provided in Appendix~\ref{app:eval-details-qa}.

\paragraph{Evaluation Protocol and Metrics.}
Following~\citet{xu2025collab,shao2023enhancing,trivedi2023interleaving}, we sample the first 500 examples from the development set of each dataset as our evaluation subset to reduce the evaluation overhead while maintaining statistical reliability. For each question, the model receives the query and its associated context passages and produces a corresponding answer. We report accuracy (\%) with respect to the GT reference answers judged by a Qwen3-235B-A22B, using the official evaluation scripts when available. We also draw {eight} independent generations for each test query and then adopt a majority voting to decide the final verdict.

\begin{table}[!h]
\centering
\caption{Comparison of MARCH-Best against RAG baselines. {MARCH-Best} denotes our best-performing model. \underline{Underline} indicates the best performance among 8B-class models, and \textbf{Bold} indicates the best overall performance. Baseline results are cited from~\citet{xu2025collab}.}
\label{tab:sota_comparison}
\resizebox{0.7\textwidth}{!}{
\begin{tabular}{l c c c c}
\toprule
{Method} & {Backbone} & {HotpotQA} & {MuSiQue} & {2WikiMQA} \\
\midrule
\rowcolor{mygray} \multicolumn{5}{c}{\textit{Advanced RAG Frameworks with Proprietary Models}} \\
CoT \citep{wei2022chain} & GPT-4o & 56.2 & 26.8 & 58.4 \\
RAG-Standard \citep{lewis2020retrieval} & GPT-4o & 64.0 & 29.8 & 57.8 \\
IRCoT \citep{trivedi2022musique} & GPT-4o & 66.4 & \textbf{44.2} & \textbf{78.0} \\
Iter-RetGen \citep{shao2023enhancing} & GPT-4o & 62.6 & 42.0 & 71.4 \\
RAG-Star \citep{jiang2024ragstarenhancingdeliberativereasoning} & GPT-4o & 57.0 & 40.0 & 63.0 \\
RQ-RAG \citep{chan2024rqraglearningrefinequeries} & GPT-4o & 62.0 & 31.6 & 60.2 \\
\midrule
\rowcolor{mygray} \multicolumn{5}{c}{\textit{Specialized RAG Micro-tuning (White-box 7B/8B)}} \\
Baseline & Llama3.1-8B & 35.0  & 5.6 & 17.4 \\
GenGround \citep{shi2024generate} & 7B-class & 47.2 & 20.2 & 45.6 \\
RAFT \citep{dong2023raftrewardrankedfinetuning} & Llama3.1-8B & 51.2 & 22.0 & 44.6 \\
IRCoT \citep{trivedi2022musique} & Llama3.1-8B & 64.0 & 29.2 & 56.8 \\
\midrule
\rowcolor{blue!5} {MARCH (10-Shots)} & {Llama3.1-8B} & \underline{\textbf{{73.6}}} & \underline{40.8} & {69.4} \\
\rowcolor{blue!5} {MARCH (CoT)} & {Llama3.1-8B} & {{70.6}} & {36.2} & \underline{}{70.6} \\
\bottomrule
\end{tabular}
}
\end{table}

\paragraph{Comparison with Existing Baselines.}
Table~\ref{tab:sota_comparison} compares MARCH-Best, which incorporates joint optimization of both the {Solver} and the {Checker} alongside CoT prompting, with various specialized RAG methods and proprietary models. Within the 8B-parameter scale, MARCH achieves higher accuracy on the HotpotQA and MuSiQue benchmarks compared to methods such as RAFT~\citep{dong2023raftrewardrankedfinetuning} and GenGround~\citep{shi2024generate}. Notably, on the HotpotQA benchmark, which contains a high density of noisy documents, MARCH reaches an accuracy of 71.2\%, exceeding the reported results of standard GPT-4o RAG (64.0\%) and the IRCoT~\citep{trivedi2023interleaving} framework using GPT-4o (66.4\%). 
Table~\ref{tab:sota_comparison} suggests that MARCH can effectively enhance the factual reliability of smaller models in multi-hop reasoning scenarios, leading to performance that is comparable to and even higher than much larger proprietary models.

\subsection{Ablation Study and Further Investigation}
\paragraph{Performance Analysis of Joint Optimization.} Table~\ref{tab:res_main_QA} illustrates that the joint optimization of the {Solver} and the {Checker} consistently achieves the highest accuracy across all training strategies in both STEM and General benchmarks. A key observation is that while the Solver-only update provides a base improvement, the inclusion of the {Checker} significantly amplifies the model's reasoning capabilities, particularly in complex scenarios. Specifically, in the \textit{Few-Shot} setting for STEM, the joint model achieves a remarkable absolute gain of 11.6\% over the {Solver-only} version. These results suggest that the {Checker}'s auditing signal is crucial for leveraging the full potential of advanced prompting techniques like \textit{few-shot} and \textit{CoT}, as it effectively mitigates factual drift and ensures logical consistency throughout multi-hop reasoning trajectories. In summary, Table~\ref{tab:res_main_QA} suggests that by internalizing a claim-level auditing logic, MARCH significantly amplifies the efficacy of in-context learning and multi-step reasoning, ensuring that increased model capacity translates into verifiable factual accuracy rather than hallucinated complexity.

\paragraph{Does our method conflict with other boosting methods?}
As demonstrated in Table~\ref{tab:res_main_QA}, MARCH is highly orthogonal and can be seamlessly integrated with existing performance-boosting techniques, including RLHF, Few-Shot prompting, and CoT reasoning. When applied to the Llama3.1-8B-Instruct baseline, our method consistently achieves significant performance gains across all benchmarks. For instance, combining MARCH with CoT reasoning further elevates the average accuracy on STEM tasks from 50.93\% to 59.13\%, and on General tasks from 51.00\% to 57.80\%. Similarly, notable improvements are observed when MARCH is integrated with Few-Shot (10-Shots) configurations, reaching an average of 61.27\% on STEM. Table~\ref{tab:res_main_QA} indicates that MARCH does not conflict with traditional scaling or prompting methods; rather, it provides a complementary layer of factual rigor through its asymmetric self-play mechanism, leading to a cumulative enhancement in multi-hop question-answering capabilities. We provide the implementation details in Appendix~\ref{app:baselines}.

\begin{table*}[!t]
\centering
\caption{Main results on Multi-Hop Question Answering benchmarks across two training datasets based on Llama3.1-8B-Instruct. We report the accuracy rate (\%) by comparing the model output with the corresponding ground-truth label. We produce baseline results on our own environment.}
\resizebox{0.99\textwidth}{!}{
\begin{tabular}{ccc|cccc|cccc}
\toprule
\multicolumn{3}{c|}{Training Dataset} & \multicolumn{4}{c|}{STEM} & \multicolumn{4}{c}{General} \\
Method & Solver & Checker & MuSiQue & 2Wiki & Hotpot & Average & MuSiQue & 2Wiki & Hotpot & Average \\ \midrule
\rowcolor{mygray} \multicolumn{11}{c}{\textit{Compare with Instruct Baseline}}   \\
\multicolumn{3}{c|}{Baseline} & 5.60 & 17.40 & 35.00 & 19.33 & 5.60 & 17.40 & 35.00 & 19.33 \\
\multirow{2}{*}{\textbf{+MARCH}} & \ding{52} & - &
12.80\textsuperscript{\scriptsize\textcolor{ggg}{$\uparrow 7.20$}} &
25.40\textsuperscript{\scriptsize\textcolor{ggg}{$\uparrow 8.00$}} &
49.40\textsuperscript{\scriptsize\textcolor{ggg}{$\uparrow 14.40$}} &
29.20\textsuperscript{\scriptsize\textcolor{ggg}{$\uparrow 9.87$}} &
12.20\textsuperscript{\scriptsize\textcolor{ggg}{$\uparrow 6.60$}} &
27.80\textsuperscript{\scriptsize\textcolor{ggg}{$\uparrow 10.40$}} &
54.80\textsuperscript{\scriptsize\textcolor{ggg}{$\uparrow 19.80$}} &
31.60\textsuperscript{\scriptsize\textcolor{ggg}{$\uparrow 12.27$}} \\
& \ding{52} & \ding{52} &
16.60\textsuperscript{\scriptsize\textcolor{ggg}{$\uparrow 3.80$}} &
35.40\textsuperscript{\scriptsize\textcolor{ggg}{$\uparrow 10.00$}} &
55.20\textsuperscript{\scriptsize\textcolor{ggg}{$\uparrow 5.80$}} &
35.73\textsuperscript{\scriptsize\textcolor{ggg}{$\uparrow 6.53$}} &
18.40\textsuperscript{\scriptsize\textcolor{ggg}{$\uparrow 6.20$}} &
38.60\textsuperscript{\scriptsize\textcolor{ggg}{$\uparrow 10.80$}} &
56.40\textsuperscript{\scriptsize\textcolor{ggg}{$\uparrow 1.60$}} &
37.80\textsuperscript{\scriptsize\textcolor{ggg}{$\uparrow 6.20$}} \\ \midrule

\rowcolor{mygray} \multicolumn{11}{c}{\textit{Compare with RLHF Baseline}}   \\
\multicolumn{3}{c|}{RLHF} & 24.20 & 49.60 & 60.40 & 44.73 & 24.20 & 56.40 & 65.40 & 48.67 \\
\multirow{2}{*}{\textbf{+MARCH}} & \ding{52} & - &
29.00\textsuperscript{\scriptsize\textcolor{ggg}{$\uparrow 4.80$}} &
49.60\textsuperscript{\scriptsize\textcolor{ggg}{$\uparrow 0.00$}} &
66.40\textsuperscript{\scriptsize\textcolor{ggg}{$\uparrow 6.00$}} &
48.33\textsuperscript{\scriptsize\textcolor{ggg}{$\uparrow 3.60$}} &
30.80\textsuperscript{\scriptsize\textcolor{ggg}{$\uparrow 6.60$}} &
56.40\textsuperscript{\scriptsize\textcolor{ggg}{$\uparrow 0.00$}} &
67.80\textsuperscript{\scriptsize\textcolor{ggg}{$\uparrow 2.40$}} &
51.67\textsuperscript{\scriptsize\textcolor{ggg}{$\uparrow 3.00$}} \\
& \ding{52} & \ding{52} &
29.40\textsuperscript{\scriptsize\textcolor{ggg}{$\uparrow 0.40$}} &
53.20\textsuperscript{\scriptsize\textcolor{ggg}{$\uparrow 3.60$}} &
66.20\textsuperscript{\scriptsize\textcolor{rrr}{$\downarrow 0.20$}} &
49.60\textsuperscript{\scriptsize\textcolor{ggg}{$\uparrow 1.27$}} &
30.80\textsuperscript{\scriptsize\textcolor{ggg}{$\uparrow 0.00$}} &
61.00\textsuperscript{\scriptsize\textcolor{ggg}{$\uparrow 4.60$}} &
69.00\textsuperscript{\scriptsize\textcolor{ggg}{$\uparrow 1.20$}} &
53.60\textsuperscript{\scriptsize\textcolor{ggg}{$\uparrow 1.93$}} \\ \midrule

\rowcolor{mygray} \multicolumn{11}{c}{\textit{Boosting with Few Shots (10-Shots)}}   \\
\multicolumn{3}{c|}{RLHF + Few Shots} & 29.20 & 55.20 & 61.20 & 48.53 & 29.40 & 55.20 & 60.60 & 48.43 \\
\multirow{2}{*}{\textbf{+MARCH}} & \ding{52} & - &
28.60\textsuperscript{\scriptsize\textcolor{rrr}{$\downarrow 0.60$}} &
58.60\textsuperscript{\scriptsize\textcolor{ggg}{$\uparrow 3.40$}} &
61.80\textsuperscript{\scriptsize\textcolor{ggg}{$\uparrow 0.60$}} &
49.67\textsuperscript{\scriptsize\textcolor{ggg}{$\uparrow 1.14$}} &
28.00\textsuperscript{\scriptsize\textcolor{rrr}{$\downarrow 1.40$}} &
57.20\textsuperscript{\scriptsize\textcolor{ggg}{$\uparrow 2.20$}} &
62.20\textsuperscript{\scriptsize\textcolor{ggg}{$\uparrow 1.60$}} &
49.13\textsuperscript{\scriptsize\textcolor{ggg}{$\uparrow 0.70$}} \\
& \ding{52} & \ding{52} &
40.80\textsuperscript{\scriptsize\textcolor{ggg}{$\uparrow 12.80$}} &
69.40\textsuperscript{\scriptsize\textcolor{ggg}{$\uparrow 10.80$}} &
73.60\textsuperscript{\scriptsize\textcolor{ggg}{$\uparrow 11.80$}} &
61.27\textsuperscript{\scriptsize\textcolor{ggg}{$\uparrow 10.60$}} &
37.20\textsuperscript{\scriptsize\textcolor{ggg}{$\uparrow 9.20$}} &
62.00\textsuperscript{\scriptsize\textcolor{ggg}{$\uparrow 4.80$}} &
64.00\textsuperscript{\scriptsize\textcolor{ggg}{$\uparrow 1.80$}} &
54.40\textsuperscript{\scriptsize\textcolor{ggg}{$\uparrow 5.27$}} \\ \midrule

\rowcolor{mygray} \multicolumn{11}{c}{\textit{Boosting with Chain-of-Thought (CoT)}}   \\
\multicolumn{3}{c|}{RLHF + CoT} & 29.40 & 59.00 & 64.40 & 50.93 &  31.60 & 58.60 & 62.80 & 51.00 \\
\multirow{2}{*}{\textbf{+MARCH}} & \ding{52} & - &
32.40\textsuperscript{\scriptsize\textcolor{ggg}{$\uparrow 3.00$}} &
60.60\textsuperscript{\scriptsize\textcolor{ggg}{$\uparrow 1.60$}} &
64.00\textsuperscript{\scriptsize\textcolor{rrr}{$\downarrow 0.40$}} &
52.33\textsuperscript{\scriptsize\textcolor{ggg}{$\uparrow 1.40$}} &
32.40\textsuperscript{\scriptsize\textcolor{ggg}{$\uparrow 0.80$}} &
62.00\textsuperscript{\scriptsize\textcolor{ggg}{$\uparrow 3.40$}} &
68.20\textsuperscript{\scriptsize\textcolor{ggg}{$\uparrow 5.40$}} &
54.20\textsuperscript{\scriptsize\textcolor{ggg}{$\uparrow 3.20$}} \\
& \ding{52} & \ding{52} &
36.20\textsuperscript{\scriptsize\textcolor{ggg}{$\uparrow 3.80$}} &
70.60\textsuperscript{\scriptsize\textcolor{ggg}{$\uparrow 10.00$}} &
70.60\textsuperscript{\scriptsize\textcolor{ggg}{$\uparrow 6.60$}} &
59.13\textsuperscript{\scriptsize\textcolor{ggg}{$\uparrow 6.80$}} &
38.40\textsuperscript{\scriptsize\textcolor{ggg}{$\uparrow 6.00$}} &
63.80\textsuperscript{\scriptsize\textcolor{ggg}{$\uparrow 1.80$}} &
71.20\textsuperscript{\scriptsize\textcolor{ggg}{$\uparrow 3.00$}} &
57.80\textsuperscript{\scriptsize\textcolor{ggg}{$\uparrow 3.60$}} \\ 
\bottomrule
\end{tabular}
}\label{tab:res_main_QA}
\end{table*}

\paragraph{Influence of Reward Formulations.} We evaluate two reward formulations: (i) {Error Rate Reward (ERR)}, a proportional penalty based on the ratio of incorrect claims, which is defined as $R_{\text{ERR}} = - \frac{N_{\text{err}}}{N_{\text{total}}}$, where $N_{\text{total}}$ is the total number of claims generated and $N_{\text{err}}$ is the number of claims identified as incorrect by the Checker. and (ii) {Zero-Tolerance Reward (ZTR)}, the binary constraint that enforces absolute consistency defined in \eqref{eq:ztr_reward}. As shown in Table~\ref{tab:all_rf_ablation}, while all RL-tuned configurations significantly outperform the vanilla baselines, ZTR consistently surpasses ERR (e.g., 61.25\% vs 55.46\% on Llama3.1 joint-training), suggesting that strict logical consistency is more effective for complex RAG tasks. Within the ZTR, the penalty-based scalar ($-1/0$) yields superior results compared to the incentive-based ($0/1$) setting, as the former provides a stronger corrective gradient to suppress hallucinations. 

\paragraph{Impact of Reward Scalars}
We further compare two scalar assignments within the ZTR framework: a penalty-based setting ($-1/0$) and an incentive-based setting ($0/1$). The second group in Table~\ref{tab:all_rf_ablation} on the STEM dataset show that the $-1/0$ configuration (59.06\%) significantly outperforms the $0/1$ setting (50.42\%). We hypothesize that because factual successes are relatively sparse during early training phases, the $0/1$ incentive fails to provide sufficient contrast between diverse erroneous paths. In contrast, the $-1/0$ setting establishes factual accuracy as the default expectation, providing a robust corrective gradient that effectively suppresses hallucinations.

\paragraph{Performance on Different Model Family.} As shown in the last group in Table~\ref{tab:all_rf_ablation}, the MARCH framework exhibits robust generalizability across different model families, with Qwen3-8B reflecting the same performance trajectory observed in Llama-based experiments. Specifically, MARCH elevates Qwen3-8B to average accuracies of 67.90\% and 68.11\% under General and STEM settings, respectively, marking a significant absolute gain of over 11\% compared to its vanilla baseline (56.84\%) and demonstrating that the dual-trajectory optimization mechanism is model-agnostic and consistently internalizes superior evidentiary grounding across diverse architectures.

\begin{table*}[t]
\centering
\caption{Ablation study on reward functions, scalars and base LLMs.  We compare Error Rate Reward (ERR) and Zero-Tolerance Reward (ZTR) under different training configurations.}
\label{tab:all_rf_ablation}
\resizebox{\textwidth}{!}{
\begin{tabular}{c|c|cc|cccc|ccc|c}
\toprule
 Reward & Reward & \multirow{2}{*}{Solver} & \multirow{2}{*}{Checker} & \multicolumn{3}{c}{RAGTruth} & \multirow{2}{*}{Facts Grounding} & \multirow{2}{*}{MuSiQue} & \multirow{2}{*}{2Wiki} & \multirow{2}{*}{Hotpot} & \multirow{2}{*}{Average} \\ 
Function & Scalar &  &  & Summary & Data2Txt & QA &  &  &  &  &  \\ \midrule
\multicolumn{4}{c|}{Llama3.1-8B Baseline} & 71.33 & 28.67 & 63.31 & 57.09 & 5.60 & 17.40 & 35.00 & 39.77 \\ 
\rowcolor{mygray} \multicolumn{12}{c}{\textit{Ablation Study on Reward Function, General Training Dataset}} \\
ERR & - & \ding{52} & - & 80.00 & 49.33 & 74.82 & 71.98 & 10.60 & 24.80 & 52.00 & 51.93 \\ 
ZTR & -1/0 & \ding{52} & - & 88.67 & 66.67 & 82.73 & 80.12 & 12.20 & 27.80 & 54.80 & 59.00 \\ \midrule
ERR & - & \ding{52} & \ding{52} & 87.33 & 62.67 & 76.26 & 76.98 & 9.80 & 25.60 & 49.60 & 55.46 \\
ZTR & -1/0 & \ding{52} & \ding{52} & 89.33 & 66.67 & 82.73 & 76.63 & 18.40 & 38.60 & 56.40 & 61.25 \\
\rowcolor{mygray} \multicolumn{12}{c}{\textit{Ablation Study on Reward Scalar, STEM Training Dataset}} \\
\multirow{4}{*}{ZTR}  & 0/1 & \ding{52} & - & 78.00 & 42.00 & 75.54 & 68.84 & 7.4 & 25.00 & 48.85 & 49.38 \\ 
& -1/0 & \ding{52} & - & 80.58 & 51.33 & 80.50 & 77.09 & 12.80 & 25.40 & 49.40 & 53.87 \\ \cmidrule{2-12}
 & 0/1 & \ding{52} & \ding{52} & 74.67 & 54.67 & 79.14 & 69.65 & 6.8 & 25.20 & 42.80 & 50.42 \\
 & -1/0 & \ding{52} & \ding{52} & 83.33 & 58.67 & 82.73 & 81.51 & 16.60 & 35.40 & 55.20 &  59.06\\ \midrule
\rowcolor{mygray} \multicolumn{12}{c}{\textit{Ablation Study on Different Base LLM}} \\
 \multicolumn{4}{c|}{Qwen3-8B Baseline} & 54.00 & 28.50 & 61.15 & 52.41 & 48.00 & 72.40 & 81.50 & 56.84 \\
\multirow{1}{*}{General} & \multirow{1}{*}{ZTR}& \ding{52} & \ding{52} & 72.00 & 47.67 & 75.90 & 64.45 & 53.20 & 76.90 & 85.20 & 67.90 \\
\multirow{1}{*}{STEM} & \multirow{1}{*}{ZTR}& \ding{52} & \ding{52} & 74.67 & 48.00 & 74.10 & 64.19 & 53.60 & 76.20 & 86.00 & 68.11 \\
 \bottomrule
\end{tabular}
}
\end{table*}

\subsection{Additional Analysis}
We analyze the potential reward hacking in MARCH and provide a primary solution (Appendix~\ref{app:question_number_varying}). We visualize the PPO training dynamics and analyze the time consumption (Appendix~\ref{app:training_dy}), where we find the consistently increasing rewards and acceptable trade-off between training consumption and improved performance. We present cases to show outputs of each agent during one training step (Appendix~\ref{app:case_study} ).

\section{Related Work}
\paragraph{Hallucinations and Fact-Check in RAG.}
While RAG reduces fabrications by anchoring responses in external context \citep{lewis2020retrieval}, LLMs remain susceptible to context-conflicting hallucinations \citep{shuster2021retrievalaugmentationreduceshallucination}. Current mitigation efforts largely revolve around outcome-level supervision~\citep{shi2023replugretrievalaugmentedblackboxlanguage,gekhman2024doesfinetuningllmsnew,yu2024chainofnoteenhancingrobustnessretrievalaugmented,asai2023selfraglearningretrievegenerate}, post-hoc reasoning refinements and utilizing external verifiers to filter outputs during inference \citep{dhuliawala2023chainofverificationreduceshallucinationlarge,madaan2023selfrefineiterativerefinementselffeedback,min-etal-2023-factscore,rashkin2023measuring,gao-etal-2023-rarr,tang2024minicheck}. However, these methods often treat LLM as a black box and fail to internalize factual constraints within the training objective itself. Moving beyond, we propose a framework where the model learns to prioritize factual adherence by minimizing logical discrepancies identified through an internal audit.

\paragraph{Multi-Agent Systems for LLMs.}
Multi-agent systems have emerged as a robust paradigm for enhancing LLM performance through collaborative or adversarial interactions \citep{du2023improvingfactualityreasoninglanguage,li2023camel}, which typically assign distinct personas to one / multiple model instances to facilitate code generation~\citep{huang2023agentcoder,islam2024mapcoder}, complex reasoning~\citep{li2023camel,he2023lego,talebirad2023multi}, decision-making~\citep{shinn2023reflexion,yu2024fincon}, recommendation~\citep{wang2024macrec}, and debate-driven consensus \citep{du2023improvingfactualityreasoninglanguage,liang2024encouraging}. While effective at inference time, most existing architectures treat agents as separate entities, often overlooking the potential for joint policy optimization across different functional roles. In addition, previous work shows that multi-agent system sometimes amplifies the hallucination of LLMs~\citep{hong2023metagpt}. Our framework differs by instantiating functionally distinct roles within a single shared policy. By structurally enforcing an information barrier between agents, we facilitate the co-evolution of generative and evaluative capabilities within a unified training loop, ensuring that the model’s internal verification logic directly informs its generative precision.

\section{Conclusion}
\vspace{-2mm}

In this paper, we introduced MARCH, a multi-agent framework designed to mitigate hallucinations in data-intensive RAG tasks. By reconfiguring a shared policy into a coordinated hierarchy comprising a Solver, a Proposer, and a Checker, MARCH enables LLMs to simultaneously refine their synthesis reasoning and independent grounding capabilities. Our experiments demonstrate that the combination of a Zero-Tolerance Reward mechanism and dual-trajectory PPO effectively internalizes a self-contained verification loop, consistently outperforming strong baselines. The robust generalization of MARCH across diverse model families underscores its potential as a scalable, high-fidelity solution for information synthesis in complex knowledge domains. We believe this agentic co-evolution paradigm offers a promising pathway toward building inherently trustworthy and verifiable large language models.

\bibliography{iclr2026_conference}
\bibliographystyle{iclr2026_conference}

\appendix

\section{Methodology}
\subsection{System Prompt of the \textit{Solver} Agent}\label{app:answer_prompt}

\begin{tcolorbox}[
    enhanced,
    colback=promptpurple,          
    colframe=bordergray,           
    boxrule=1pt,                   
    sharp corners,                 
    fontupper=\ttfamily,           
    left=15pt,                     
    right=15pt,                    
    top=12pt,                      
    bottom=12pt                    
]
\begin{flushleft}
    \textbf{Instruction:} \\
    You must respond based strictly on the information in provided passages. 
    Do not incorporate any external knowledge or infer any details beyond what is given in the passages. \\
    \vspace{1em}
    
    \textbf{Passages $D$:} \\
    \{Document-1\} \\
    ... \\
    \{Document-K\} \\
    \vspace{1em}
    
    \textbf{Query $\mathbf{x}$:} \\
    \{User Query $\mathbf{x}$\} \\
    \vspace{1em}
    
    \textbf{Output:}
\end{flushleft}
\end{tcolorbox}

\subsection{System Prompt of the \textit{Proposer} Agent}\label{app:propose_qa}
\begin{tcolorbox}[
    enhanced,
    colback=promptgreen,           
    colframe=bordergray,           
    boxrule=1pt,                   
    sharp corners,                 
    fontupper=\ttfamily,           
    left=15pt,                     
    right=15pt,                    
    top=12pt,                      
    bottom=12pt                    
]
\begin{flushleft}
    \textbf{You are a data annotation expert.} Your task is to analyze a model's response to a user query. For every critical number found in the model's response, you must create a question where the answer is precisely and exactly that number. \\
    
    \textbf{Rule 1. Clarity and Specificity of Questions:} Each question must be clear, complete, and unambiguous. It must contain enough context from the response to ensure the number is the only possible correct answer. For example, if the model's response contains ``In 2025, 50 people will take the bar exam in Beijing,'' an effective question would be ``How many people will take the bar exam in the Beijing area in 2025?'', and the answer is 50. A bad example would be ``How many people will take the bar exam in 2025?''. This is ambiguous because it lacks the specific region ('Beijing area'). \\
    
    \textbf{Rule 2. Format of the Answer:} The answer must be a pure number, either an integer or a decimal. It must NOT include any other characters or formats, such as percentage signs (\%), 'k' to denote thousands (e.g., 10k), ranges (e.g., 10-20), or words (e.g., fifty). \\
    
    Please provide your output as an unordered list using the following format for each question-answer pair: \\
    \texttt{- Question: xxx [Answer: n]} \\
    \texttt{- Question: xxx [Answer: n]} \\
    
    Now, generate questions for the user query and model response below. Do not output any other content or explanations. \\
    
    \vspace{1em}
    
    \textbf{Model Response:} \\
    \{response $\mathbf{y}$\}
\end{flushleft}
\end{tcolorbox}

\subsection{System Prompt of the \textit{Checker} Agent}\label{app:check_qa}
\begin{tcolorbox}[
    enhanced,
    breakable,
    colback=promptblue,            
    colframe=bordergray,           
    boxrule=1pt,                   
    sharp corners,                 
    fontupper=\ttfamily,           
    left=15pt,                     
    right=15pt,                    
    top=12pt,                      
    bottom=12pt                    
]
\begin{flushleft}
    \textbf{You are a data validation expert.} You will be given a set of reference materials and a list of user questions. Your task is to retrieve information from the reference materials to answer all questions. Before providing each answer, you must state the evidence for it. \\
    
    You must adhere strictly to the content of the reference materials and must not fabricate or infer any information that is not explicitly mentioned. If the relevant information cannot be found in the materials, you must state ``Cannot answer''. It is important to note that all answers are pure numbers. Therefore, your answer must also be a pure number and must not contain any other content, such as percentage signs (\%), 'k' to denote thousands, ranges, or numbers written as words. \\
    
    \textit{Here is an example:} \\
    \textbf{Reference Materials:} \\
    Document 1: ...in 2024, 50 people will take the bar exam in Beijing. \\
    Document 2: ...(omitted) \\
    
    \vspace{0.5em}
    - Question 1: How many people will take the bar exam in Beijing in 2024? \\
    - Question 2: How many people will take the bar exam in Beijing in 2025? \\
    
    \vspace{0.5em}
    \textbf{Example Response:} \\
    1. Evidence: Document 1 states that 50 people will take the bar exam in Beijing in 2024.\\
    \texttt{[Answer: 50]} \\
    2. Evidence: The materials do not contain information about the bar exam in Beijing for 2025, therefore I cannot answer. \\
    \texttt{[Answer: Cannot answer]} \\
    
    Now, based on the reference materials below, answer all the following questions. Strictly adhere to the example format provided. \\
    
    \vspace{1em}
    \textbf{Reference Materials:} \\
    \{docs $D$\} \\
    
    \textbf{Questions:} \\
    \{questions $\mathbf{q}^{\text{aux}}$\}
\end{flushleft}
\end{tcolorbox}

\section{Experiment}\label{app:exp}

\subsection{Training Details}\label{app:training_hp}
We give more detailed training hyper-parameters in the Table~\ref{tab:hyperparams}.

\begin{table}[!h]
\centering
\caption{Key hyperparameters for our RL training, derived from the execution script.}
\label{tab:hyperparams}
\resizebox{0.7\textwidth}{!}{%
\begin{tabular}{llc}
\toprule
\multirow{3}{*}{Data \& Model Configuration} & Max Prompt Length & 24,567 \\
& Max Response Length & 8,192 \\
& Filter Overlong Prompts & True \\
\midrule
\multirow{4}{*}{Optimizer \& Learning Rates} & Actor Learning Rate & \(1 \times 10^{-6}\) \\
& Critic Learning Rate & \(1 \times 10^{-5}\) \\
& Actor LR Warmup Steps & 5 \\
& Critic LR Warmup Steps & 10 \\
\midrule
\multirow{2}{*}{PPO Algorithm \& Rewards} & $\gamma$ & 0.998 \\
& KL Divergence Coefficient & \(1 \times 10^{-3}\) \\
\midrule
\multirow{2}{*}{Training Schedule} & Total Epochs & 1 \\
& Global Batch Size / Mini-batch Size & 32 \\
\midrule
\multirow{2}{*}{Rollout (Training)} & Inference Engine & \texttt{vLLM} \\
&  Temperature & 0.6 \\ \midrule
\multirow{6}{*}{Rollout (Evaluation)} & Inference Engine & \texttt{vLLM} \\
&  Temperature & 0.6 \\
&  Batch Size & 32/32 \\
&  Top-P & 0.95 \\
&  Top-K & 20 \\
&  Rollout Number & 8 \\
\bottomrule
\end{tabular}%
}
\end{table}
\subsection{Evaluation Details}

\subsubsection{Hallucination and Factual Grounding Benchmarks}\label{app:eval-details-fc}
\paragraph{RAGTruth and FaithBench} are public benchmarks for evaluating hallucinations in retrieval-augmented generation, which contain three task types: (i) QA, (ii) Data2Txt Writing, and (iii) Summarization, each with human-annotated hallucinations at the response level. For each example, the model is given a query and a set of retrieved passages, and asked to produce an answer, a data-to-text description, or a summary. 

\paragraph{Contextual JudgeBench} is a judge benchmark with 2,000 challenging response pairs across eight splits inspired by real-world contextual evaluation scenarios, which is built with a multi-pronged data construction pipeline that leverages both existing human annotations and model-based perturbations. The evaluation task requires a model to act as a judge, choosing the better of two provided responses (A and B) according to specific, context-dependent criteria.

\paragraph{Facts Grounding} is a large-scale factuality benchmark spanning five domains (e.g., medical, financial, legal, and others) and nine task types, including summarization, fact finding, comparison, and definition. Each instance consists of a query, a collection of documents, and a reference output with fine-grained factual annotations, providing a broad and challenging testbed for document-grounded evaluation. During test, we adopt the official \texttt{JSON} judgment prompt provided by~\citet{jacovi2025facts}.

\subsubsection{QA Benchmarks}\label{app:eval-details-qa}

\paragraph{HotpotQA.}
HotpotQA~\citep{yang2018hotpotqa} is a large-scale multi-hop QA dataset that requires reasoning over multiple Wikipedia passages to answer fact-based questions. Each question is paired with multiple supporting paragraphs and a short, factoid-style answer.

\paragraph{2WikiMultiHopQA.}
2WikiMultiHopQA~\citep{ho2020constructing} extends HotpotQA with questions that require reasoning across two different Wikipedia articles, emphasizing cross-document retrieval and multi-hop reasoning over disjoint evidence.

\paragraph{MuSiQue.}
MuSiQue~\citep{trivedi2022musique} is a challenging multi-hop QA dataset with questions that explicitly require reasoning across multiple sentences scattered across different documents, often involving compositional reasoning over several intermediate facts.

\subsection{Baseline Methods}
\label{app:baselines}

\paragraph{Baselines.}
The raw backbone model used in all settings, evaluated in a zero-shot manner with task-specific instructions but without additional reasoning or RL fine-tuning. 

\paragraph{+CoT.}
A prompting baseline where we encourage step-by-step reasoning by prepending chain-of-thought instructions (e.g., ``let's think step by step'') to the prompt. The model is still the same base LLM; only the decoding prompt is changed.

\paragraph{+Few-Shot.}
An in-context learning baseline where we augment the prompt with a small number of annotated QA examples (few-shot demonstrations) drawn from the training distribution, following standard practice in multi-hop QA and factuality evaluation.

\paragraph{+RLHF.}
A version of the base LLM further optimized with reinforcement learning from human feedback, using general-purpose reward models that favor helpful and harmless responses but are not specifically tuned for document-grounded factuality. We adopt Skywork-Reward-V2-Llama-3.1-8B~\citep{liu2025skyworkrewardv2scalingpreferencedata} as the reward model.

\subsection{How Would Proposed Question Numbers Influence Performance?}\label{app:question_number_varying}
\begin{figure}[!h]
    \centering
    \includegraphics[width=0.5\linewidth]{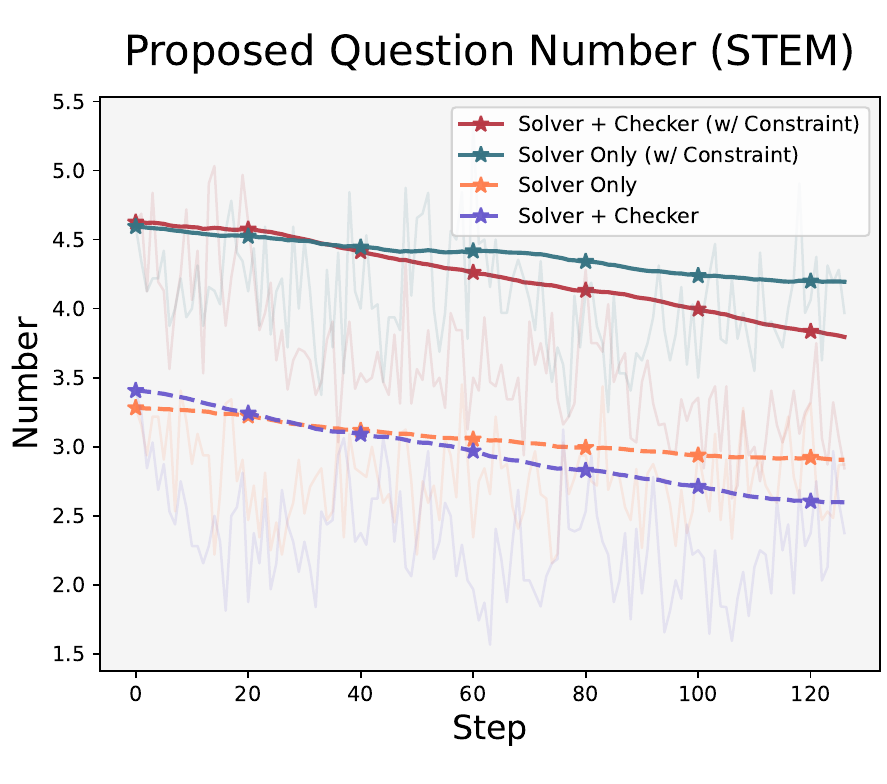}
    \caption{Average number of proposed questions per step on the STEM training dataset. The dashed lines represent the MARCH framework without maintaining the question number, while the solid lines (w/ Constraint) indicate the results after incorporating instructional constraints to maintain informational density.}
    \label{fig:q_number_stem}
\end{figure}

A potential reward hacking of the MARCH framework could be that the system may attempt to inflate its reward by reducing the number of proposed questions, adhering to a ``less said, less leaked'' strategy where fewer claims minimize the probability of being audited as a hallucination. As illustrated in Figure~\ref{fig:q_number_stem}, we indeed observe a gradual decline in the average number of proposed questions as training progresses across all configurations. Notably, the configuration involving joint updates of the {Solver + Checker} exhibits a more pronounced downward trend compared to the {Solver Only} baseline, suggesting that the zero-tolerance penalty on the Checker's trajectory imposes a stricter consistency bottleneck, further incentivizing the \textit{Proposer} to prune potentially risky claims.

However, we find that this quantitative shrinkage can be effectively mitigated through simple instructional constraints. By explicitly requiring the \textit{Proposer} to generate no fewer than $k$ questions within the prompt (e.g., $k=3$), the decline in question density is significantly arrested. As shown by the solid lines in Figure~\ref{fig:q_number_stem}, models with such constraints in the legend (solid lines marked with ``w/ constraint'') maintain a much more stable and higher volume of factual inquiries throughout the self-evolution process, which indicates that the MARCH framework remains robust when balanced with basic task-specific instructions, ensuring that the drive for factual accuracy does not come at the cost of informational density.

\subsection{Visualization of Training Dynamics and Time Consuming}\label{app:training_dy}
\begin{figure*}[!h]
    \centering
    \includegraphics[width=1\linewidth]{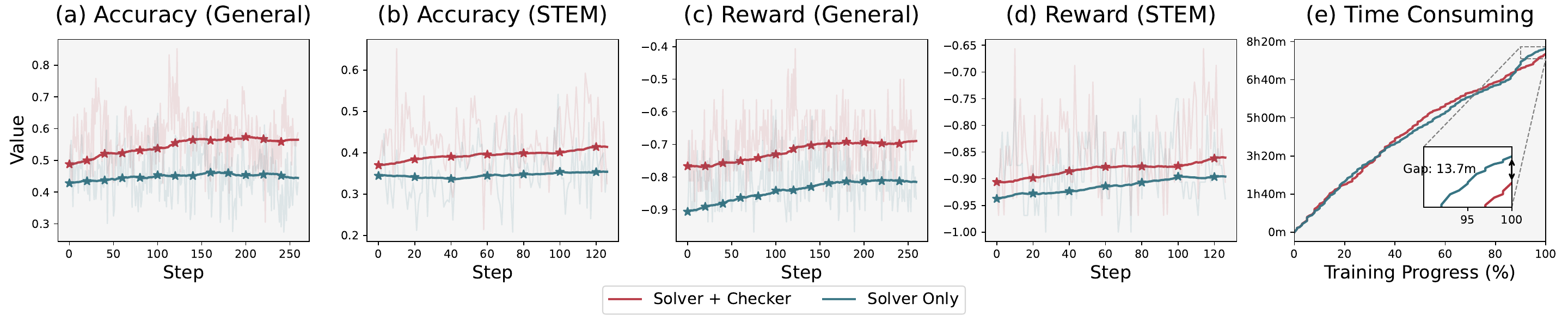}
\caption{Visualization of training efficiency and dynamics. (a)-(d) illustrate the convergence of accuracy and rewards across general and STEM datasets. (e) Comparison of cumulative training time on the General dataset.}
    \label{fig:training_dyncs}
\end{figure*}

To further investigate the practical efficiency of the MARCH framework, we analyze the training dynamics and computational overhead. Figures~\ref{fig:training_dyncs}(a)-(d) illustrate the stable convergence of accuracy and rewards across both General and STEM datasets, where the joint {Solver + Checker} update consistently outperforms the {Solver Only} configuration. Regarding training efficiency, Figure~\ref{fig:training_dyncs}(e) presents the cumulative time consumption on the General dataset (containing 9,000 samples), where we find that simultaneously updating both the \textit{Solver} and the \textit{Checker} trajectories in a single training step does not impose a prohibitive computational burden.

\section{Case Study}\label{app:case_study}
\subsection{Case 1}
\paragraph{Input $(\vx,\mD)$ of Agent Solver $\nu_\text{solve}$}

\begin{tcolorbox}[
    enhanced,
    breakable,
    colback=promptpurple,            
    colframe=bordergray,           
    boxrule=1pt,                   
    sharp corners,                 
]
\begin{flushleft}
You must respond based strictly on the information in the passages. Do not incorporate any external knowledge or infer any details beyond what is given in the passages.
\\
\vspace{1em}
\textbf{Passages:}\\
\textbf{Document-1}\\
TUDY OBJECTIVE: To describe sleep characteristics and rapid eye movement (REM) sleep behavior disorder in patients with Guadeloupean atypical parkinsonism (Gd-PSP)...(omit)...
Subjects with idiopathic REM sleep behavior disorder should avoid eating sour sop.
\\
\vspace{1em}
\textbf{Document-2}\\
We characterize the clinical features of Parkinson's syndrome on Guadeloupe and describe possible environmental causes...(omit)...
The development of atypical parkinsonism in Guadeloupe and probably elsewhere, could result from synergistic toxicity, but acetogenins are probably the most potent neurotoxin, acting as mitochondrial complex I inhibitor.
\\
\vspace{1em}
\textbf{Document-3}\\
In Guadeloupe, there is an abnormally high frequency of atypical parkinsonism...(omit)...
In conclusion, atypical Guadeloupean parkinsonism comprises two forms of parkinsonism and dementia that differ clinically by the presence of oculomotor signs, but have similar cognitive profiles and neuroimaging features, suggesting that they may constitute a single disease entity, and both were similarly exposed to annonaceous neurotoxins, notably annonacin.
\\
\vspace{1em}
\textbf{Document-4}\\
An unusually high frequency of atypical Parkinson syndrome has been delineated over the last 5 years in the French West Indies...(omit)... 
Guadeloupean parkinsonism may prove to be a tauopathy identical or closely related to PSP.
\\
\vspace{1em}
\textbf{Document-5}\\
INTRODUCTION: On Guadeloupe, atypical parkinsonism is abnormally frequent, and represents 75\% of progressive parkinsonism while Parkinson's disease (PD) accounts for only 25\%, which is an inversed percentage in comparison with Europe...(omit)...
This hypothesis is in keeping with epidemiologic data and experimental studies showing neuronal loss after exposure to isoquinolines or acetogenins.
\\
\vspace{1em}
\textbf{Document-6}
\\
Steele, Richardson and Olszweski in 1964 described a distinctive clinical and pathological entity they called progressive supranuclear palsy (PSP)...(omit)...
Future studies will confirm if Guadeloupean Parkinsonism is truly a geographic focus of PSP, and if dietary factors account for both.
\\
\vspace{1em}
\textbf{Document-7}
\\
On the French West Indian island of Guadeloupe...(omit)...
Consistent with this view, chronic administration of annonacin to rats through Alzet osmotic minipumps showed that annonacin was able to reproduce the brain lesions characteristic of the human disease.
\\
\vspace{1em}
\textbf{Document-8}
\\
Atypical parkinsonism is extremely frequent in Guadeloupe and may have an environmental cause...(omit)..
This electrophysiological study, added to previous clinical, neuropsychological and neuroradiological studies, has enriched the characterization of Guadeloupean atypical parkinsonism, which thus appears to be a new clinical entity.
\\
\vspace{1em}
\textbf{Question:}\\
Describe clinical presentation of Parkinsonism with dementia of Guadeloupe syndrome.
\end{flushleft}
\end{tcolorbox}

\paragraph{Generated Response $\vy$ of Agent Solver $\nu_\text{solve}$ by $\vy\sim\nu_\text{solve}(\cdot|\vx,\mD)$}
\begin{tcolorbox}[
    enhanced,
    breakable,
    colback=promptpurple,            
    colframe=bordergray,           
    boxrule=1pt,                   
    sharp corners,                 
]
\begin{flushleft}
The clinical presentation of Parkinsonism with dementia in Guadeloupe syndrome includes the following features:  \\n1. **Levodopa Resistance**: Patients exhibit levodopa-resistant parkinsonism, distinguishing them from typical Parkinson's disease.  \\n2. **Fronto-Subcortical Dementia**: Cognitive deficits are primarily fronto-subcortical, characterized by frontolimbic cognitive impairment, hallucinations (52–59\%), and dementia with features such as dysautonomia (50\%) and tremor (>50\%).  \\n3. **Oculomotor Abnormalities**: Some patients display supranuclear oculomotor dysfunction, including reduced vertical saccade velocity and impaired antisaccades. However, oculomotor dysfunction is not universal, as seen in the parkinsonism-dementia complex (PDC) subgroup.  \\n4. **Tremor and Myoclonus**: Jerky tremor and cortical myoclonus (89\% of cases) are prominent, with myoclonus originating from cortical regions.  \\n5. **REM Sleep Behavior Disorder (RBD)**: 78\% of patients with Guadeloupean atypical parkinsonism (Gd-PSP) experience RBD, often preceding parkinsonism symptoms by years.  \\n6. **Neuroimaging Findings**: Cerebral atrophy is common, with enlargement of the third ventricle and T2-hypointensity in the basal ganglia (particularly substantia nigra). Neuronal loss affects the substantia nigra, striatum, and frontal cortex, alongside tau-containing lesions.  \\n7. **Environmental Factor**: Higher consumption of soursop (Annona muricata), a plant containing the mitochondrial complex I inhibitor annonacin, is significantly associated with the syndrome.  \\nThese features collectively define Guadeloupe syndrome as a distinct clinical entity characterized by parkinsonism, dementia, and unique neuroimaging and environmental correlates.
\end{flushleft}
\end{tcolorbox}

\paragraph{Atomize $\vy$ into verifiable claims through Agent Proposer $\nu_\text{propose}$}

\begin{tcolorbox}[
    enhanced,
    breakable,
    colback=promptgreen,           
    colframe=bordergray,           
    boxrule=1pt,                   
    sharp corners,                 
]
\begin{flushleft}
    \textbf{You are a data annotation expert.} Your task is to analyze a model's response to a user query. For every critical number found in the model's response, you must create a question where the answer is precisely and exactly that number. \\
    
    \textbf{Rule 1. Clarity and Specificity of Questions:} Each question must be clear, complete, and unambiguous. It must contain enough context from the response to ensure the number is the only possible correct answer. For example, if the model's response contains ``In 2025, 50 people will take the bar exam in Beijing,'' an effective question would be ``How many people will take the bar exam in the Beijing area in 2025?'', and the answer is 50. A bad example would be ``How many people will take the bar exam in 2025?''. This is ambiguous because it lacks the specific region ('Beijing area'). \\
    
    \textbf{Rule 2. Format of the Answer:} The answer must be a pure number, either an integer or a decimal. It must NOT include any other characters or formats, such as percentage signs (\%), 'k' to denote thousands (e.g., 10k), ranges (e.g., 10-20), or words (e.g., fifty). \\
    
    Please provide your output as an unordered list using the following format for each question-answer pair: \\
    \texttt{- Question: xxx [Answer: n]} \\
    \texttt{- Question: xxx [Answer: n]} \\
    
    Now, generate questions for the user query and model response below. Do not output any other content or explanations. \\
    
    \vspace{1em}    
    \textbf{Model Response:} 
    
    The clinical presentation of Parkinsonism with dementia in Guadeloupe syndrome includes the following features:  \\n1. **Levodopa Resistance**: Patients exhibit levodopa-resistant parkinsonism, distinguishing them from typical Parkinson's disease.  \\n2. **Fronto-Subcortical Dementia**: Cognitive deficits are primarily fronto-subcortical, characterized by frontolimbic cognitive impairment, hallucinations (52–59\%), and dementia with features such as dysautonomia (50\%) and tremor (>50\%).  \\n3. **Oculomotor Abnormalities**: Some patients display supranuclear oculomotor dysfunction, including reduced vertical saccade velocity and impaired antisaccades. However, oculomotor dysfunction is not universal, as seen in the parkinsonism-dementia complex (PDC) subgroup.  \\n4. **Tremor and Myoclonus**: Jerky tremor and cortical myoclonus (89\% of cases) are prominent, with myoclonus originating from cortical regions.  \\n5. **REM Sleep Behavior Disorder (RBD)**: 78\% of patients with Guadeloupean atypical parkinsonism (Gd-PSP) experience RBD, often preceding parkinsonism symptoms by years.  \\n6. **Neuroimaging Findings**: Cerebral atrophy is common, with enlargement of the third ventricle and T2-hypointensity in the basal ganglia (particularly substantia nigra). Neuronal loss affects the substantia nigra, striatum, and frontal cortex, alongside tau-containing lesions.  \\n7. **Environmental Factor**: Higher consumption of soursop (Annona muricata), a plant containing the mitochondrial complex I inhibitor annonacin, is significantly associated with the syndrome.  \\nThese features collectively define Guadeloupe syndrome as a distinct clinical entity characterized by parkinsonism, dementia, and unique neuroimaging and environmental correlates.
    
\end{flushleft}
\end{tcolorbox}

\paragraph{Generated Question-Answers pairs of the \textit{Proposer}.}

\begin{tcolorbox}[
    enhanced,
    breakable,
    colback=promptgreen,           
    colframe=bordergray,           
    boxrule=1pt,                   
    sharp corners,                 
]
\begin{flushleft}
    - Question: What is the lower percentage range of patients with Guadeloupe syndrome experiencing hallucinations?
    [Answer: 52]
    \vspace{1em}
    
    - Question: What is the upper percentage range of patients with Guadeloupe syndrome experiencing hallucinations?
    [Answer: 59]
    \vspace{1em}
    
    - Question: What percentage of patients with Guadeloupe syndrome have dysautonomia?
    [Answer: 50]
    \vspace{1em}
    
    - Question: What percentage of patients with Guadeloupe syndrome have cortical myoclonus?
    [Answer: 89]
    \vspace{1em}
    
    - Question: What percentage of patients with Guadeloupe syndrome experience REM Sleep Behavior Disorder (RBD)?
    \vspace{1em}
    [Answer: 78]
\end{flushleft}
\end{tcolorbox}

\paragraph{Input $(\{\vq_i\}_{i=1}^n,\mD)$ of Agent Checker $\nu_\text{check}$}

\begin{tcolorbox}[
    enhanced,
    breakable,
    colback=promptblue,            
    colframe=bordergray,           
    boxrule=1pt,                   
    sharp corners,                 
]
\begin{flushleft}
    \textbf{You are a data validation expert.} You will be given a set of reference materials and a list of user questions. Your task is to retrieve information from the reference materials to answer all questions. Before providing each answer, you must state the evidence for it. \\
    
    You must adhere strictly to the content of the reference materials and must not fabricate or infer any information that is not explicitly mentioned. If the relevant information cannot be found in the materials, you must state ``Cannot answer''. It is important to note that all answers are pure numbers. Therefore, your answer must also be a pure number and must not contain any other content, such as percentage signs (\%), 'k' to denote thousands, ranges, or numbers written as words. \\
    
    \textit{Here is an example:} \\
    \textbf{Reference Materials:} \\
    Document 1: ...in 2024, 50 people will take the bar exam in Beijing. \\
    Document 2: ...(omitted) \\
    
    \vspace{0.5em}
    - Question 1: How many people will take the bar exam in Beijing in 2024? \\
    - Question 2: How many people will take the bar exam in Beijing in 2025? \\
    
    \vspace{0.5em}
    \textbf{Example Response:} \\
    1. Evidence: Document 1 states that 50 people will take the bar exam in Beijing in 2024.\\
    \texttt{[Answer: 50]} \\
    2. Evidence: The materials do not contain information about the bar exam in Beijing for 2025, therefore I cannot answer. \\
    \texttt{[Answer: Cannot answer]} \\
    
    Now, based on the reference materials below, answer all the following questions. Strictly adhere to the example format provided. \\
    
    \vspace{1em}
    \textbf{Reference Materials:} 
    \\
    \vspace{1em}
    \textbf{Document-1} (omit)\\
    \vspace{1em}
    \textbf{Document-2} (omit)\\
    \vspace{1em}
    \textbf{Document-3} (omit)\\
    \vspace{1em}
    \textbf{Document-4} (omit)\\
    \vspace{1em}
    \textbf{Document-5} (omit)\\
    \vspace{1em}
    \textbf{Document-6} (omit)\\
    \vspace{1em}
    \textbf{Document-7} (omit)\\
    \vspace{1em}
    \textbf{Document-8} (omit)\\
    \vspace{1em}
    
    \textbf{Questions:} \\
    - Question: What is the lower percentage range of patients with Guadeloupe syndrome experiencing hallucinations?
    \\
    \vspace{1em}
    - Question: What is the upper percentage range of patients with Guadeloupe syndrome experiencing hallucinations?
    \\
    \vspace{1em}
    - Question: What percentage of patients with Guadeloupe syndrome have dysautonomia?
    \\
    \vspace{1em}
    - Question: What percentage of patients with Guadeloupe syndrome have cortical myoclonus?
    \\
    \vspace{1em}
    - Question: What percentage of patients with Guadeloupe syndrome experience REM Sleep Behavior Disorder (RBD)?
    
\end{flushleft}
\end{tcolorbox}

\paragraph{Generated Response $\{\hat{\va}_i\}_{i=1}^n$ of Agent Checker $\nu_\text{check}$ by $\{\hat{\va}_i\}_{i=1}^n\sim\nu_\text{check}(\cdot|\{{\vq}_i\}_{i=1}^n,\mD)$}

\begin{tcolorbox}[
    enhanced,
    breakable,
    colback=promptblue,            
    colframe=bordergray,           
    boxrule=1pt,                   
    sharp corners,                 
]
\begin{flushleft}
1. Evidence: Document-5 states that 52\% of PDC patients had hallucinations, which is the lower percentage range mentioned for Guadeloupe syndrome.\\
Answer: 52
\vspace{1em}

2. Evidence: Document-3 states that 59\% of patients with PSP-like syndrome experienced hallucinations, which is the upper percentage range mentioned for Guadeloupe syndrome.\\
Answer: 59
\vspace{1em}

3. Evidence: Document-3 states that dysautonomia was present in 50\% of patients with Guadeloupe syndrome.\\
Answer: 50 
\vspace{1em}

4. Evidence: Document-8 states that 89\% of Gd-PSP patients had cortical myoclonus.\\
Answer: 89 
\vspace{1em}

5. Evidence: Document-1 states that 78\% of patients with Gd-PSP experienced REM sleep behavior disorder.\\
Answer: 78 
\vspace{1em}
\end{flushleft}
\end{tcolorbox}

\paragraph{Calculate Zero-Tolerance Reward (ZTR) according to $\{\va_i\}_{i=1}^n$ contained in $\vy$ from Agent Solver $\nu_\text{solve}$ and $\{\hat{\va}_i\}_{i=1}^n$ generated by Agent Checker $\nu_\text{check}$.}
\end{document}